\documentclass[10pt,twocolumn,letterpaper]{article}

\usepackage{iccv}              % To produce the CAMERA-READY version
\usepackage{makecell}
\usepackage{colortbl}
\usepackage{multicol}
\usepackage{multirow}
\usepackage{diagbox}
\usepackage{mdframed}
\usepackage{tikz}
\usepackage{adjustbox}
\usepackage[most]{tcolorbox}
\tcbuselibrary{skins}

\newtcolorbox{mybox}[1][]{enhanced,
    colback=gray!10,
    colframe=black,
    boxrule=1pt,
    arc=5pt,
    auto outer arc,
    boxsep=5pt,
    left=3pt,
    right=0pt,
    top=5pt,
    bottom=0pt,
    breakable,
    sharp corners=all,
    #1
}
\newcommand{\bench}{ICE-Bench }

\definecolor{iccvblue}{rgb}{0.21,0.49,0.74}
\usepackage[pagebackref,breaklinks,colorlinks,allcolors=iccvblue]{hyperref}

\title{ICE-Bench: A Unified and Comprehensive Benchmark for \\Image Creating and Editing}

\author{
Yulin Pan$^1$ \quad
Xiangteng He$^2$ \quad
Chaojie Mao$^1$ \quad
Zhen Han$^1$ \\
Zeyinzi Jiang$^1$ \quad
Jingfeng Zhang$^1$ \quad
Yu Liu$^1$ 
\\[5pt]
$^1$Tongyi Lab, Alibaba Group \quad
$^2$Peking University
}

\let\oldtwocolumn\twocolumn
\renewcommand\twocolumn[1]
[]{
    \oldtwocolumn[{#1}{
    \begin{center} \includegraphics[width=\textwidth]{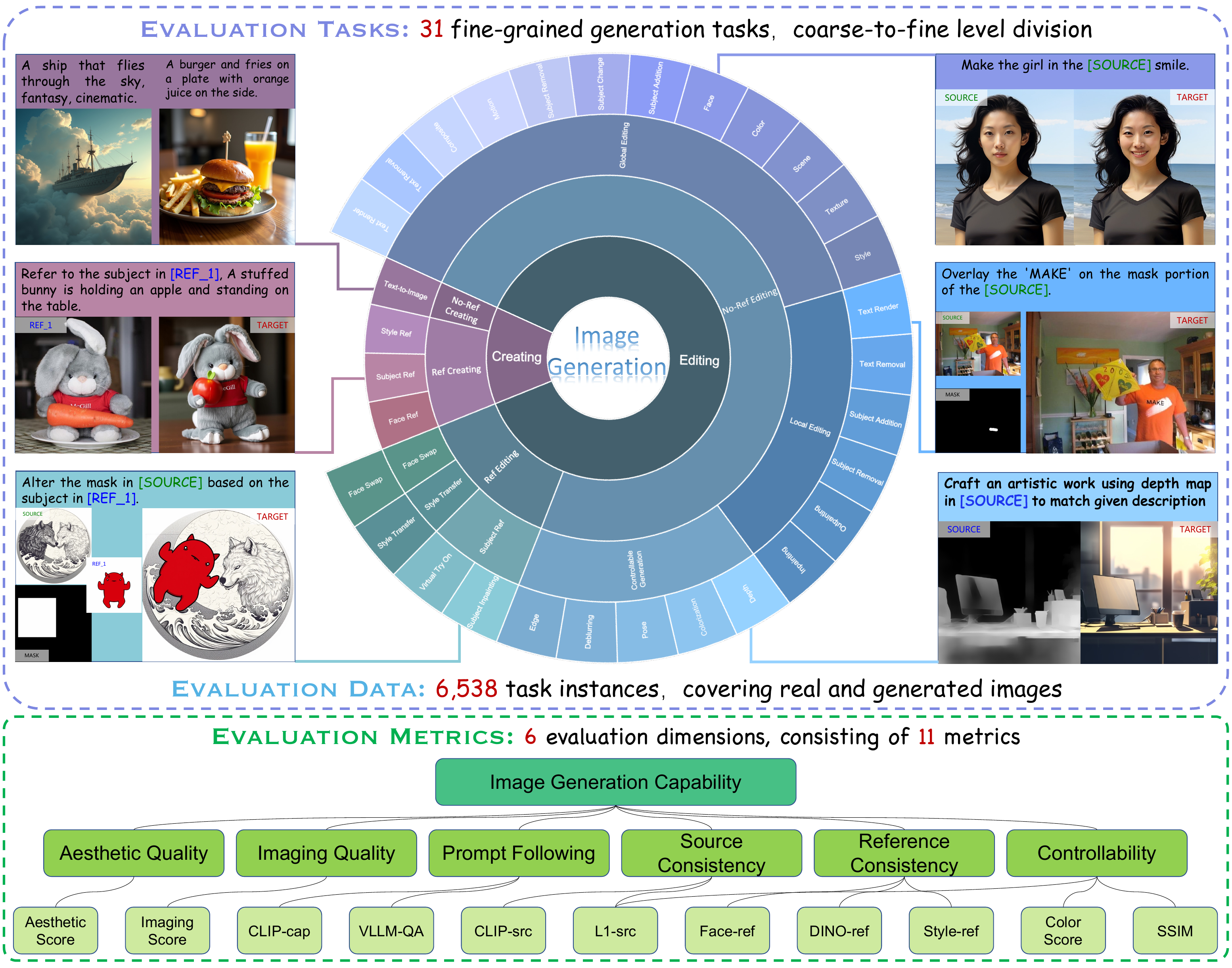}
        \captionof{figure}{Overview of our \bench, including evaluation tasks, data, and metrics. }
        \label{fig:teaser}
    \end{center}
    }]
}

\begin{document}
\maketitle
\begin{abstract}
Image generation has witnessed significant advancements in the past few years. However, evaluating the performance of image generation models remains a formidable challenge. 
In this paper, we propose \textbf{\bench}, a unified and comprehensive benchmark designed to rigorously assess image generation models. Its comprehensiveness could be summarized in the following key features:
(1) \textbf{Coarse-to-Fine Tasks}: We systematically deconstruct image generation into four task categories: No-ref/Ref Image Creating/Editing, based on the presence or absence of source images and reference images. And further decompose them into 31 fine-grained tasks covering a broad spectrum of image generation requirements, culminating in a comprehensive benchmark.
(2) \textbf{Multi-dimensional Metrics}: The evaluation framework assesses image generation capabilities across 6 dimensions: aesthetic quality, imaging quality, prompt following, source consistency, reference consistency, and controllability. 11 metrics are introduced to support the multi-dimensional evaluation. 
Notably, we introduce VLLM-QA, an innovative metric designed to assess the success of image editing by leveraging large models.
(3) \textbf{Hybrid Data}: The data comes from real scenes and virtual generation, which effectively improves data diversity and alleviates the bias problem in model evaluation. 
Through \bench, we conduct a thorough analysis of existing generation models, revealing both the challenging nature of our benchmark and the gap between current model capabilities and real-world generation requirements.
To foster further advancements in the field, we will open-source \bench, including its dataset, evaluation code, and models, thereby providing a valuable resource for the research community. The detail of \bench can be found at \url{https://ali-vilab.github.io/ICE-Bench-Page/}
\end{abstract}    
\section{Introduction}
\label{sec:intro}

\begin{table*}[ht]
  \centering
   \caption{Comparison with existing image generation datasets.}
  \resizebox{\textwidth}{!}{
      \begin{tabular}{@{}lccccccc@{}}
        \toprule
        Datasets & Publications & Creating  & Editing & Training & Evaluation &\#Evaluation Data & \#Evaluation Tasks \\
        \midrule
        CUB-200-2011 ~\cite{CUB-2011,CUB-2011-2} & \makecell{Caltech techreport 2011, \\ CVPR 2016} & \checkmark &  & \checkmark & \checkmark & 2,933 & 1 \\
        Oxford Flower-102 ~\cite{OxfordFlower} & ICVGIP 2008 & \checkmark &  & \checkmark & \checkmark & 2,040 & 1 \\
        MS-COCO ~\cite{MSCOCO} & ECCV 2014 & \checkmark &  & \checkmark & \checkmark & 40,000 & 1 \\
        DrawBench ~\cite{DrawBench} & NeurIPS 2022 & \checkmark &  &  & \checkmark(Human-eval) & 200 & 1 \\
        Multi-Task Benchmark ~\cite{MultiTaskBench} & NeurIPSW 2022 & \checkmark &  &  & \checkmark(Human-eval) & 3,600 & 1 \\
        PAINTSKILLS ~\cite{PaintSkills} & ICCV 2023 & \checkmark &  &  & \checkmark & 7,185 & 1 \\
        EditBench ~\cite{EditBench} & CVPR 2023 &  & \checkmark &  & \checkmark & 240 & 1 \\
        InstructPix2Pix ~\cite{IP2P} & CVPR 2023 &  & \checkmark & \checkmark &  & - & 4 \\
        MagicBrush ~\cite{MagicBrush} & NeurIPS 2023 &  & \checkmark & \checkmark & \checkmark & 1,053 & 5 \\
        UltraEdit ~\cite{UltraEdit} & NeurIPS 2024 &  & \checkmark & \checkmark & & - & 9+ \\
        HIVE ~\cite{Hive} & CVPR 2024 &  & \checkmark & \checkmark & \checkmark & 1,000 & 1 \\
        Emu Edit~\cite{EmuEdit} & CVPR 2024 & & \checkmark &  & \checkmark & 3,055 & 7 \\
        %HQ-Edit ~\cite{hui2024hq} & ICLR 2025 &  & \checkmark & \checkmark & & - & 6 \\
        \midrule
        \rowcolor{gray!20}
        \textbf{ICE-Bench} & \textbf{This paper} & \textbf{\checkmark} & \textbf{\checkmark}  &  & \textbf{\checkmark}  & \textbf{6,538} & \textbf{31}\\
        \bottomrule
      \end{tabular}
  }
  \vspace{-8pt}
  \label{tab:compdataset}
\end{table*}

Image generation has witnessed remarkable advancements in recent years, driven by significant technological breakthroughs such as VAEs~\cite{VAE, VAR, Infinity}, GANs~\cite{GAN}, and Diffusion Models~\cite{DDPM, rf, lcm, ldm}.
Image generation encompasses a broad array of tasks aimed at producing high-quality images that satisfy multiple criteria, including aesthetic appeal, imaging quality, and adherence to the given description, among others.
Image generation tasks are boundless, owing to the wide range of conditions and the intricate nature of natural language-based instructions.

Existing image generation foundational models~\cite{sd15, sd21, sdxl, sd3, DALLE-3, FLUX, dit} predominantly focus on text-to-image creating, which aims to generate images based on textual descriptions. When addressing more complex tasks, most existing methods~\cite{LARGen, anydoor, SDEdit, controlnet, scedit, pbe, sdinp, OminiControl} opt for minor architecture adjustments to the text-to-image foundational model, followed by parameter fine-tuning to tackle the specific task.
Recently, efforts have been made to develop unified architectures capable of handling comprehensive image generation tasks, as exemplified by models like ACE~\cite{ACE} and OmniGen~\cite{OmniGen}. Research on unified image generation foundation models is garnering increasing attention as it reduces the semantic gap between the pre-trained models and practical applications, and significantly lowers the costs associated with task-specific customization.

Despite this growing research trend, the development of automatic evaluation benchmarks for unified image generation remains significantly lagging.
Early evaluation frameworks primarily relied on datasets such as CUB-200-2011~\cite{CUB-2011,CUB-2011-2}, Oxford Flower-102\cite{OxfordFlower}, and MS-COCO~\cite{MSCOCO}, and utilized Fr\'{e}chet Inception Distance (FID)~\cite{FID} and Inception Score (IS)~\cite{IS} as metrics to quantify the performance. 
However, these data and metrics are  primarily tailored for text-to-image creating, limiting their ability to assess global description-guided generation, could not reflect the comprehensive capability of a unified image generation model.
%
% EditBench~\cite{EditBench} is proposed for the evaluation of text-guided image inpainting, and MagicBrush~\cite{MagicBrush} offers a more extensive dataset supporting multiple editing tasks, including single-/multi-turn editing, mask-provided, and mask-free scenarios, making it a more versatile evaluation benchmark. 
Recently some image editing benchmarks~\cite{IP2P, MagicBrush, EmuEdit} have been proposed to assess model performance on general-purpose instruction-based image editing tasks.
While these benchmarks provide valuable insights, they exhibit several limitations when used to evaluate unified image generation models:
\textit{(1) Limited Evaluation Scope.} As illustrated in \cref{tab:compdataset}, existing datasets are typically tailored for one or a few specific tasks, resulting in evaluation outcomes that are biased toward these particular tasks.
\textit{(2) Insufficient Evaluation Granularity and Dimensions:} Current evaluation frameworks predominantly rely on metrics such as FID and IS for assessing image quality, and CLIP~\cite{CLIP} similarity for image-condition consistency. However, these metrics are inadequate for comprehensive evaluation and often misalign with human preferences. As a result, human evaluation is often necessary, making the process both time-consuming and costly.
\textit{(3) Bias in Data Distribution:} Most existing benchmarks suffer from data bias issues. For example, InstructPix2Pix~\cite{IP2P} includes only synthesized images, while MagicBrush~\cite{MagicBrush} contains only real images. This limitation hampers the ability to evaluate model performance across diverse data sources.

Given these limitations, there is an urgent need for a unified and comprehensive benchmark to evaluate image generation models both automatically and effectively. 
Therefore, we propose \textbf{\bench}, an extensive benchmark designed for the holistic evaluation of unified image generation. 
As demonstrated in \cref{tab:compdataset}, the strengths of our \bench can be encapsulated in three key aspects: \textit{coarse-to-fine tasks}, \textit{multi-dimensional metrics}, and \textit{hybrid data}.

First, we establish a \textbf{hierarchical evaluation task set} that decomposes image generation capabilities into coarse-to-fine granularity. 
As shown in \cref{fig:teaser}, the evaluation tasks are categorized into 2 dimensions at a coarse-grained dimensions based on generation type (creating and editing). Each of them is further divided into 2 medium-grained categories based on dependency on a reference (Ref or Non-Ref).
Then 31 specific tasks are broken down into more refined categories, considering controllable generation, global editing, local editing, style transferring, etc. Based on these 31 coarse-to-fine evaluation tasks, the model's generation ability can be comprehensively evaluated. 

Second, in conjunction with the evaluation tasks, we have planned \textbf{6 evaluation dimensions} to effectively evaluate the model capabilities, including aesthetic quality, imaging quality, prompt following, source consistency, reference consistency, and controllability.
These dimensions are quantified using 11 specialized metrics, providing targeted insights into model performance and guiding improvements in model architecture, training data, and strategies.

Third, we curate a \textbf{hybrid dataset} that includes both real-world and virtually generated images.
In this way, data diversity is effectively promoted, and the bias problem in evaluation is alleviated.

We conduct extensive experiments and evaluate the generation capability of 10 state-of-the-art generation models, including both general-purpose and task-specific models. 
The experimental results highlight their strengths and weaknesses across various tasks and dimensions. We are open-sourcing \bench, including data, evaluation script, and models, to encourage broader participation in advancing image generation research and fostering fair, comprehensive, and unified evaluation practices.

\section{Related work}
\label{sec:related}

%In \cref{tab:compdataset}, various image generation datasets are compared from the aspects of evaluation task types and data size. Based on their evaluation task type, we split them into 2 categories, i.e., text-to-image creating and image editing.

\subsection{Text-to-image Creating Benchmark}
In the early stages, CUB-200-2011~\cite{CUB-2011,CUB-2011-2}, Oxford Flower-102~\cite{OxfordFlower}, and MS-COCO~\cite{MSCOCO} were commonly used to evaluate generation capabilities. These datasets primarily focus on a single generation task, namely text-to-image creating, and are domain-specific. For instance, CUB-200-2011 and Oxford Flower-102 are limited to specific categories such as birds and flowers, respectively. This narrow focus makes it challenging to assess the performance of generation models in real-world applications.
Moreover, these datasets primarily evaluate two aspects: image quality and instruction alignment, using metrics such as FID~\cite{FID} and CLIP score~\cite{CLIP}. To further evaluate the generation models' visual reasoning skills, PAINTSKILLS~\cite{PaintSkills} was proposed, which could measure three fundamental visual reasoning abilities, i.e., object recognition, object counting, and spatial relation understanding. To assess text-to-image models in greater depth, DrawBench~\cite{DrawBench} was proposed to assess capabilities such as rendering colors, object counts, spatial relationships, and text within scenes. However, DrawBench relies on human evaluation, which limits its generalizability, a limitation shared by the Multi-Task Benchmark~\cite{MultiTaskBench}. 

\subsection{Image Editing Benchmark}
While the aforementioned datasets are primarily designed for text-to-image creating task, image editing has become a fundamental practice in today's digital landscape, playing a crucial role in fields such as photography, advertising, and social media.
Consequently, researchers have begun to develop datasets specifically for image editing tasks.  InstructPix2Pix~\cite{IP2P} and UltraEdit~\cite{UltraEdit}
%, and HQ-Edit~\cite{hui2024hq} 
leverage LLMs such as GPT-3~\cite{GPT-3} and GPT-4~\cite{GPT-4}, to generate the image editing instructions. It is noted that these datasets are primarily used for training generation models rather than evaluation. Similarly, HIVE~\cite{Hive},  although containing evaluation data, is designed for training purposes, utilizing human feedback for instructional image editing. And it also focuses on a single editing task: generating an image based on an original image and an instruction. MagicBrush~\cite{MagicBrush} and Emu Eidt~\cite{EmuEdit} cover multiple editing tasks, i.e., 5 and 7 types, respectively.  However, these datasets still fall short of providing a comprehensive evaluation of generative models' capabilities.

Given the limitations of existing datasets, it remains challenging to evaluate the capabilities of generation models comprehensively. 
To this end, we propose \textbf{\bench}, a benchmark that includes 31 fine-grained creating and editing tasks, and 11 evaluation metrics taking account of aesthetic quality, imaging quality, prompt following, source consistency, reference consistency, and controllability. 
\section{\bench}
\label{sec:method}

In this section, we introduce \bench, our proposed benchmark for evaluating image generation models. 
We first elaborate on the overall 31 image generation tasks in \cref{subsec:evaltask}, which are categorized into 4 groups, in terms of the format of input data.
Next, we detail the evaluation metrics in \cref{subsec:evalmetric}, which are designed to comprehensively assess the generation capabilities of existing models across 6 dimensions. 
Finally, in \cref{subsec:datacons}, we introduce the overall data construction pipeline to show its reliability.
%The experiments and insights will be introduced in \cref{sec:exp} and \cref{sec:dis} respectively.

\subsection{Evaluation Tasks}
\label{subsec:evaltask}
Recall that \bench aims to comprehensively evaluate the model capability in the multi-modal guided image generation field.
To achieve this goal, we first curate and design 31 specific image generation tasks. Examples of these tasks are illustrated in Fig. 1 of Supplementary File.
Despite the diverse final objectives of these tasks, we find that only up to three types of inputs are necessary for all existing image generation tasks: 
\textit{(1) Textual prompt}, could be the user's instructions or descriptions for the generated image. 
\textit{(2) Source image}, provides to accommodate specific regional editing requests while maintaining pixel consistency in areas unrelated to the editing requests. 
\textit{(3) Reference images}, provides reference on some aspects, typically specified by the instruction.

Based on the presence or absence of a source image and reference images, 31 tasks are classified into 4 categories: (1) \textit{No-ref Image Creating}, which relies solely on a textual prompt as the condition; (2) \textit{No-ref Image Editing}, which requires both a textual prompt and a source image as input; (3) \textit{Ref Image Creating}, which necessitates a textual prompt along with reference images; and (4) \textit{Ref Image Editing}, which demands all three types of inputs for generation. 

\subsubsection{No-ref Image Creating}
\begin{itemize}[leftmargin=10pt]
\setlength{\itemsep}{2pt}
\setlength{\parsep}{0pt}
\setlength{\parskip}{0pt}
\item
\textbf{Task 1: Text-to-Image Creating}. Generate an image based solely on a given textual prompt.
\end{itemize}

\subsubsection{Ref Image Creating}
\begin{itemize}[leftmargin=10pt]
\setlength{\itemsep}{2pt}
\setlength{\parsep}{0pt}
\setlength{\parskip}{0pt}
\item
\textbf{Tasks 2-4: Face/Style/Subject Reference Creating}. These tasks require generating an image that adheres to a textual prompt while maintaining consistency with a reference image in terms of \textit{facial features, style, or subject}.
\end{itemize}

\subsubsection{No-ref Image Editing}
Here we further split the generation capabilities in disentangled 3 aspects: \textit{global editing, local editing, and controllable generation}.

\noindent\textbf{Global Editing:}
\begin{itemize}[leftmargin=10pt]
\setlength{\itemsep}{2pt}
\setlength{\parsep}{0pt}
\setlength{\parskip}{0pt}
\item \textbf{Tasks 5-8: Color/Motion/Face/Texture Editing}. 
These tasks involve modifying specific \textit{attributes (e.g., color, motion, facial features, or texture)} of a source image based on textual instructions.
\item \textbf{Task 9: Style Editing}. Change the style of a source image to match a specified style described in the  instruction. 
\item \textbf{Task 10: Scene Editing}. Following the instruction, alter the background of the source image, to a specific scene. 
\item \textbf{Tasks 11-13: Subject Addition/Removal/Change}. According to the instructions, \textit{add, delete or change an object to another object} on the given source image, while preserving other regions.
\item \textbf{Tasks 14-15: Text Render/Removal}. Render some text in source image at location specified by instruction or remove all text. 
\item \textbf{Task 16: Composite Editing}. Perform multiple edits on a source image based on instructions. 
\end{itemize}

\noindent\textbf{Local Editing}:
\begin{itemize}[leftmargin=10pt]
\setlength{\itemsep}{2pt}
\setlength{\parsep}{0pt}
\setlength{\parskip}{0pt}
\item \textbf{Task 17: Inpainting}. Repaint a specific region of the source image, identified by a mask image, with content specified by a textual description or instructions.
\item \textbf{Task 18: Outpainting}. Expand the source image beyond its original boundary based on a mask and textual prompt.
\item \textbf{Tasks 19-20: Local Subject Addition/Removal}. Like subject addition/removal, but the location is indicated by a mask. 
\item \textbf{Tasks 21-22: Local Text Render/Removal}. Like text render/removal, but the location is indicated by a mask. 
\end{itemize}

\noindent\textbf{Controllable Generation}:
\begin{itemize}[leftmargin=10pt]
\setlength{\itemsep}{2pt}
\setlength{\parsep}{0pt}
\setlength{\parskip}{0pt}
\item 
\textbf{Tasks 23-25: Pose/Edge/Depth-guided Generation}. Generate an image that aligns with the given \textit{pose/edge/depth} map.
\item
\textbf{Task 26: Image Colorization}. Given a gray image as source image, colorize it by following the textual prompt. 
\item
\textbf{Task 27: Image Deblurring}. Given a low-quality or blurred image, make it clear and improve its quality.
\end{itemize}

\subsubsection{Ref Image Editing}
\begin{itemize}[leftmargin=10pt]
\setlength{\itemsep}{2pt}
\setlength{\parsep}{0pt}
\setlength{\parskip}{0pt}
\item \textbf{Task 28: Style Transfer}. Transfer the style of a reference image to a source image.
\item \textbf{Task 29: Subject-guided Inpainting}. Incorporate a subject from a reference image into a specific region of a source image, as indicated by a mask image.
\item \textbf{Task 30: Virtual Try On}. A  specialized application of subject-guided  inpainting focused on rendering clothing items realistically on human subjects.
\item \textbf{Task 31: Face Swap}. Regenerate the face in the masked area of the source image by referencing the facial features of the reference image.
\end{itemize}

\subsection{Evaluation Metrics}
\label{subsec:evalmetric}

We assess the performance of existing generative models across 6 dimensions: \textit{aesthetic quality, imaging quality, prompt following, source consistency, reference consistency, and controllability}, as illustrated in \cref{fig:teaser}.
Here we present the metrics employed for each evaluation dimension. Detailed calculation equations can be found in the Supplementary Material.

\subsubsection{Aesthetic Quality}
We assess the \textit{aesthetic score} of generated images using the Aesthetic Predictor\footnote{https://github.com/discus0434/aesthetic-predictor-v2-5}
%LAION~\cite{LAIONAES} aesthetic predictor
, which reflects the overall aesthetic appeal of the generated image, considering multiple aspects such as layout, colorfulness, harmony, naturalness, and photo-realism.

\subsubsection{Imaging Quality}
The imaging quality assessment focuses mainly on aspects such as blur, noise, distortion, and overexposure. We utilize the MUSIQ~\cite{MUSIQ} quality predictor
%, trained on the koniq dataset, 
as the \textit{imaging score}. To ensure a fair comparison, we resize all generated images to same shape, which tends to favor high-resolution images.

\subsubsection{Prompt Following}
For image creating tasks, we measure prompt following using  the \textit{CLIP similarity} between the text prompt and the generated image. 
While, for image editing tasks, we dissect the prompt following ability into two aspects, i.e., visual-language alignment, and instruction execution reasoning. It is noted that we make the best use of the powerful ability of VLLM in these two evaluations.
Firstly, for each test case, a reasonable description of target image is inferred by providing a VLLM with all input conditions. Then we calculate the CLIP similarity between the generated image and the given caption to reflect the degree of prompt following, denoted as \textit{CLIP-cap}.
Secondly, we newly introduce a \textit{VLLM-QA} metric to assess whether an edit instruction has been successfully executed, utilizing the reasoning ability of VLLM. Given the textual instruction, source image, reference images, and the generated image, we prompt VLLM, i.e., Qwen2-VL-72B~\cite{QWEN2VL}, to determine the success of the generation. A successful generation returns a value of 1, otherwise returns 0.

\subsubsection{Source Consistency}
For image editing tasks, we evaluate the semantic consistency and pixel alignment between the source image and the generated image, based on the principle that any image editing request needs retain some pixel unchanged to some extent. We calculate CLIP similarity for semantic consistency and mean L1 distance for pixel alignment between source image and generated image, denoted as \textit{CLIP-src and L1-src}, respectively.

\subsubsection{Reference Consistency}
For reference-based tasks, the reference consistency should be considered from different dimensions in terms of the task objective. In our \bench, we mainly focus on three types of reference: face, subject and style.
For face reference, we calculate the \textit{face similarity} using the buffalo model from InsightFace App~\cite{InsightFace}. 
For subject reference, we calculate the \textit{DINO~\cite{DINO} similarity} between the reference image and generated image. 
For style reference, we use the CSD~\cite{CSD} model to extract feature as style descriptor, then calculate the similarity between reference image and generated image, denoted as \textit{Style-ref}.

\subsubsection{Controllability}
Controllable generation aims to generate an artistic image under the control of some low-level visual cues, such as edge map, depth map, pose map, \textit{etc}. We use diverse metrics that tailored for difference control conditions to comprehensively evaluate the controllability of models:
(1) For pose, depth and edge-guided generation tasks, we extract the low-level feature of generated image~\cite{canny, depth, openpose} and then calculate the mean l1 distance between the input and the extracted feature maps.
(2) For image colorization task, we calculate the colorfulness score~\cite{Colorfulness} of generated image, and then calculate the mean l1 distance between the input image and the grayscale generated image.
(3) For image deblurring task, we calculate the SSIM~\cite{SSIM} score between the source image and generated image.

\begin{figure}[t]
  \centering
  \includegraphics[width=\linewidth]{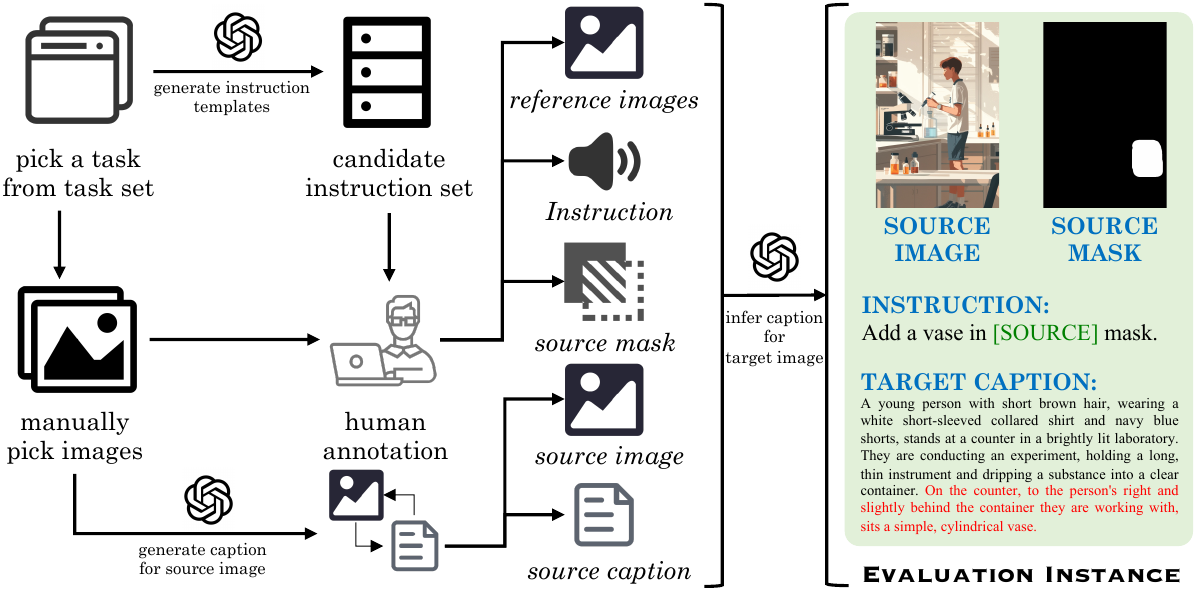}
   \caption{The overall pipeline of dataset construction.}
   \label{fig:datacons}
   \vspace{-8pt}
\end{figure}

\subsection{Dataset Construction}
\label{subsec:datacons}
\begin{figure}[t]
  \centering
  \includegraphics[width=\linewidth]{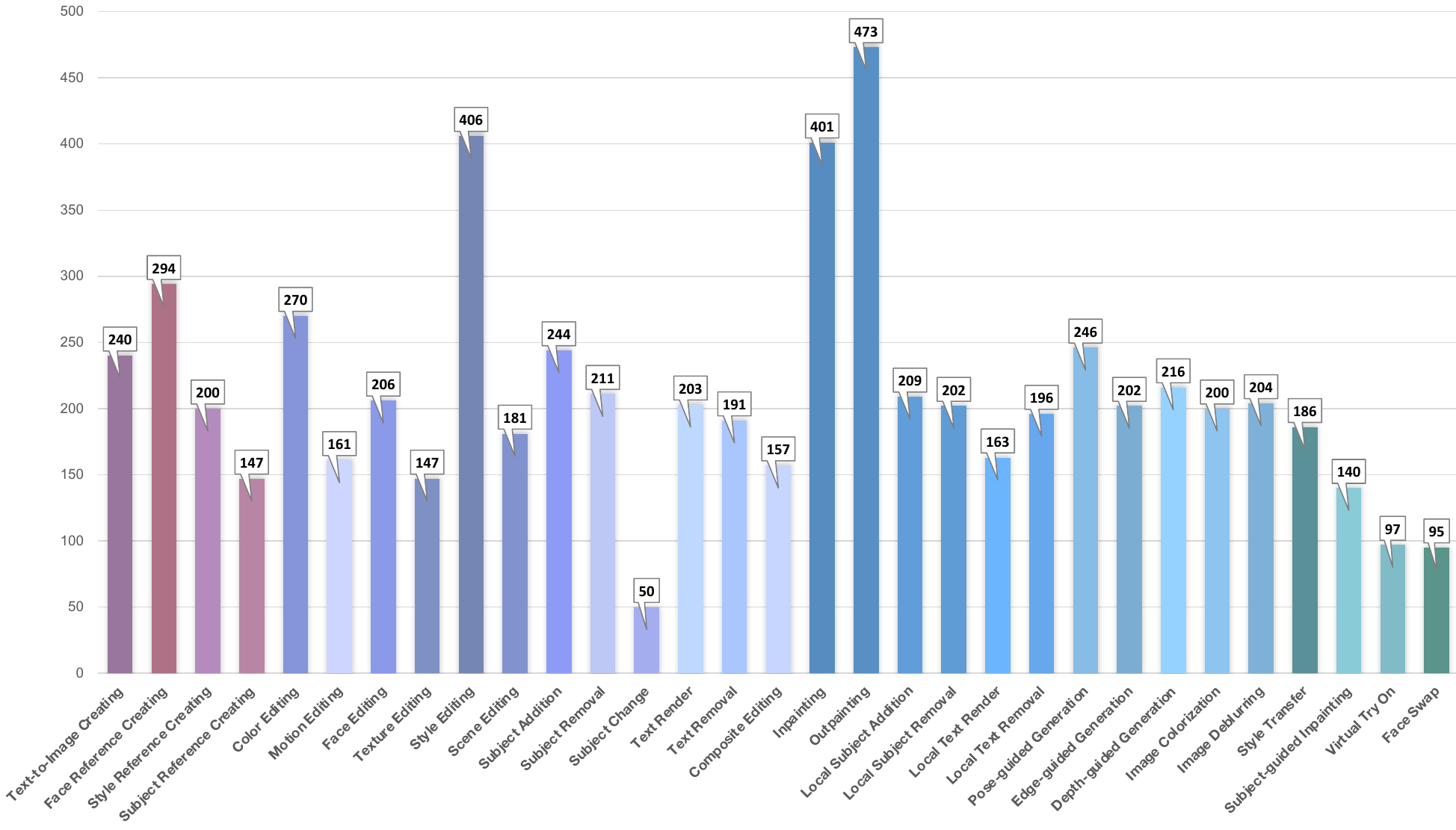}
   \caption{Data distribution of each task in \bench.}
   \label{fig:datadis}
   \vspace{-10pt}
\end{figure}

Collecting a dataset for 31 fine-grained evaluation tasks demands a significant investment of both human resources and time. 
To streamline this process, we have designed a data construction pipeline that integrates large pre-trained models with human annotation, as depicted in \cref{fig:datacons}. 
The pipeline begins with the manual selection of source and reference images from hybrid datasets such as MS-COCO~\cite{MSCOCO}, LAION-5B~\cite{laion5b}, DreamBooth~\cite{dreambooth}, VITON-HD~\cite{vitonhd} and synthetic image databases. 
Next, we utilize VLLMs to generate descriptions for the chosen source images and provide task-specific instruction templates.
Based on the instruction templates, we request annotators to craft unique instructions for each case, taking into account the selected source image and reference image. The annotation process follows rigorous standards to ensure the accuracy and reliability.
Finally, using all the available data, we ask VLLM to envision the content of an ideal generated image and to produce a detailed description of this imagined image.
In total, we collect 6,538 instances across all tasks. The distribution of the dataset is detailed in \cref{fig:datadis}.

\section{Experiments}
\label{sec:exp}

%In this section, we evaluate the generation capability of existing 10 image generation models on our \bench systematically. First, in \cref{subsec:setup}, we introduce the generation models evaluated on our \bench, and experimental settings. Then, quantitative and qualitative results are reported in \cref{subsec:results}.

%\subsection{Experiment Setup}
%\label{subsec:setup}

\subsection{Evaluation Models}

We select 10 mainstream image generation models to comprehensively evaluate their capabilities in 31 tasks from 6 diverse aspects, which have been introduced in \cref{sec:method}, including OmniGen~\cite{OmniGen}, OminiControl~\cite{OminiControl}, ACE~\cite{ACE}, ACE++~\cite{ACE++}, FLUX~\cite{FLUX}, FLUX-Control~\cite{FLUX-Control}, InstructPix2Pix ~\cite{IP2P}, MagicBrush~\cite{MagicBrush}, UltraEdit~\cite{UltraEdit}, IP-Adapter~\cite{IP-Adapter}. All evaluation metrics are designed such that higher values indicate better performance.

However, it is important to note that not all models are designed to excel in every evaluation task.  For example, FLUX is limited to text-to-image creating, while OminiControl specialize in subject reference creating and controllable generation tasks. 
To ensure a fair comparison, we first evaluate each model on its designated tasks and subsequently conduct a comprehensive cross-task analysis. For tasks that a model cannot address, its score is set to zero. Detailed task-perspective and model-perspective analysis are presented in \cref{subsec:task_spec_exp} and \cref{subsec:model_spec_exp}, respectively.

\subsection{Evaluation Setting}
All experiments are conducted using the Diffusers \footnote{https://github.com/huggingface/diffusers} framework for model implementation. For models not yet integrated into Diffusers, we utilize their official  implementations. All generation and evaluation processes are performed on A100 GPU cards, with default hyper-parameters for each model to ensure consistency and reproducibility.

\subsection{Task-perspective Evaluation and Comparison}
\label{subsec:task_spec_exp}

Given the extensive scope of evaluation tasks, we categorize the comparison into four task settings: \textit{No-ref Image Creating, Ref Image Creating, No-ref Image Editing, Ref Image Editing}, as introduced in \cref{subsec:evaltask}.
Here, we present the evaluation results for these four task subgroups, with the specific score values for each task detailed in the Supplementary Material.

\subsubsection{Evaluation on No-ref Image Creating Tasks}
\begin{table}[t]
  \centering
  \caption{\mbox{Model performance on No-ref Image Creating Task (Task 1).}}
  \vspace{-8pt}
  \small
  \begin{tabular}{c|ccc}
    \toprule
    Models & AES$\uparrow$ & IMG$\uparrow$ & PF$\uparrow$ \\
    \midrule
    ACE & 0.548  & 0.534  &  0.566 \\
    OmniGen & 0.611  & 0.726  & \textbf{0.570}  \\
    FLUX & \textbf{0.618}  & \textbf{0.735}  & \textbf{0.570}  \\
    \bottomrule
  \end{tabular}
  \vspace{-8pt}
  \label{tab:eval_noref_cre}
\end{table}
\begin{table*}[t]
  \centering
  \caption{Model performance on Ref Image Creating Tasks.}
  \vspace{-8pt}
  \small
  \resizebox{0.9\textwidth}{!}{
      \begin{tabular}{c|cccc|cccc|cccc}
        \toprule
        \multirow{2}{*}{Models} & \multicolumn{4}{c}{\textbf{Face Reference (Task 2)}} & \multicolumn{4}{c}{\textbf{Style Reference (Task 3)}} & \multicolumn{4}{c}{\textbf{Subject Reference (Task 4)}} \\
        & AES$\uparrow$ & IMG$\uparrow$ & PF$\uparrow$ & REF$\uparrow$ & AES$\uparrow$ & IMG$\uparrow$ & PF$\uparrow$ & REF$\uparrow$ & AES$\uparrow$ & IMG$\uparrow$ & PF$\uparrow$ & REF$\uparrow$ \\
        \midrule 
        ACE         & 0.535  & 0.550  & 0.531   & 0.329 & 0.531 & 0.590 & 0.232 & \textbf{0.802} & 0.523 & 0.557 & 0.498       & \textbf{0.878}     \\
        OmniGen     & \textbf{0.579}  & \textbf{0.727}  & \textbf{0.541}   & 0.573 & \textbf{0.579} & \textbf{0.708} & \textbf{0.431} & 0.432 & \textbf{0.582} & 0.714 & \textbf{0.532}       & 0.753     \\
        IP-Adapter  & 0.505  & 0.642  & 0.508   & \textbf{0.633} & 0.577 & 0.696 & 0.288 & 0.749 & 0.573 & 0.703 & 0.484       & 0.841     \\
        ACE++       & 0.551  & 0.679  & 0.523   & 0.506 & -     & -     & -     & -     & 0.520 & 0.628 & 0.475       & 0.853     \\
        OminiControl& -      & -      & -       & -     & -     & -     & -     & -     & 0.565 & \textbf{0.723} & 0.528       & 0.783     \\
        \bottomrule
      \end{tabular}
  }
  \label{tab:eval_ref_cre}
\end{table*}
\begin{table*}[t]
  \centering
  \caption{Model performance on No-Ref Image Editing Tasks. It should be noted that OminiControl lacks pose-guided generation capability, which results in its average score in Controllable Generation being significantly lower than that of other methods.}
  \vspace{-8pt}
  \small
  \resizebox{0.9\textwidth}{!}{
      \begin{tabular}{c|cccc|cccc|cccc}
        \toprule
        \multirow{2}{*}{Models} &  \multicolumn{4}{c}{\textbf{Global Editing (Tasks 5-16)}} & \multicolumn{4}{c}{\textbf{Local Editing (Tasks 17-22)}} & \multicolumn{4}{c}{\textbf{Controllable Generation (Tasks 23-27)}} \\
        & AES$\uparrow$ & IMG$\uparrow$ & PF$\uparrow$ & SRC$\uparrow$ & AES$\uparrow$ & IMG$\uparrow$ & PF$\uparrow$ & SRC$\uparrow$ & AES$\uparrow$ & IMG$\uparrow$ & PF$\uparrow$ & CTRL$\uparrow$\\
        \midrule 
        ACE         & \textbf{0.498} & 0.521 & \textbf{0.567} & \textbf{0.899} & 0.492 & 0.490 & \textbf{0.615}       & 0.919   & 0.545  & 0.505  & \textbf{0.630}   & \textbf{0.865}   \\
        OmniGen      & 0.490 & \textbf{0.598} & 0.520 & 0.844 & 0.464 & 0.562 & 0.503       & 0.853   & 0.510  & \textbf{0.594}  & 0.592   & 0.783  \\
        InstructPix2Pix  & 0.480 & 0.518 & 0.355 & 0.758 & -     & -     & -           & -   & -      & -      & -       & -       \\
        MagicBrush  & 0.456 & 0.486 & 0.444 & 0.840 & -     & -     & -           & -   & -      & -      & -       & -      \\
        UltraEdit   & 0.490 & 0.504 & 0.470 & 0.866 & 0.454 & 0.455 & 0.432       & \textbf{0.953}  & -      & -      & -       & -   \\
        FLUX-Control & -     & -     & -     & -     & -     & -     & -           & -    & \textbf{0.552}  & 0.534  & 0.609   & 0.846      \\
        ACE++  & -     & -     & -     & -     & \textbf{0.495} & \textbf{0.588} & 0.606       & 0.933 & -      & -      & -       & -    \\
        OminiControl* & -     & -     & -     & -     & -     & -     & -           & -   & 0.416  & 0.380  & 0.451   & 0.687       \\
        \bottomrule
      \end{tabular}
  }
  \label{tab:eval_noref_edit}
\end{table*}
\begin{table*}[!ht]
  \centering
  \caption{Model performance on Ref Image Editing Tasks.}
  \vspace{-8pt}
  \small
  \resizebox{0.9\textwidth}{!}{
      \begin{tabular}{c|ccccc|ccccc|ccccc}
        \toprule
        \multirow{2}{*}{Models} & \multicolumn{5}{c}{\textbf{Style Transfer (Task 28)}} & \multicolumn{5}{c}{\textbf{Subject Reference (Tasks 29-30)}} & \multicolumn{5}{c}{\textbf{Face Swap (Task 31)}} \\
        & AES$\uparrow$ & IMG$\uparrow$ & PF$\uparrow$ & REF$\uparrow$ & SRC$\uparrow$ & AES$\uparrow$ & IMG$\uparrow$ & PF$\uparrow$ & REF$\uparrow$ & SRC$\uparrow$ & AES$\uparrow$ & IMG$\uparrow$ & PF$\uparrow$ & REF$\uparrow$ & SRC$\uparrow$ \\
        \midrule 
        ACE  & \textbf{0.535} & 0.530 & \textbf{0.350} & 0.234 & \textbf{0.788} & \textbf{0.483} & 0.586  & 0.415 & 0.657 & \textbf{0.909}  & 0.498  & 0.570  & 0.432   & 0.250 & \textbf{0.873}   \\
        OmniGen     & 0.505 & \textbf{0.630} & 0.338 & \textbf{0.359} & 0.701 & 0.458 & 0.667  & 0.414 & 0.650 & 0.821  & 0.431  & 0.640  & \textbf{0.459}   & \textbf{0.477} & 0.775    \\
        ACE++  & -     & -     & -     & -     & -     & 0.471 & \textbf{0.685}  & \textbf{0.480} & \textbf{0.663} & 0.892  & \textbf{0.503}  & \textbf{0.650}  & 0.452   & 0.378 & 0.853    \\
        \bottomrule
      \end{tabular}
  }
  \vspace{-8pt}
  \label{tab:eval_ref_edit}
\end{table*}

\noindent There is only one task belonging to this group, \textit{i.e.}, text-to-image creating, and 3 models, \textit{i.e.}, ACE, OmniGen, and FLUX, favor this task. We assess their performance from 3 aspects including aesthetic quality (AES), imaging quality (IMG) and prompt following (PF), as shown in \cref{tab:eval_noref_cre}.
% %
% We observe that both FLUX.1-dev and OmniGen significantly outperform ACE, primarily due to differences in model size. ACE has a size of only 0.6 billion parameters, while OmniGen contains 3.8 billion and FLUX.1-dev boasts 12 billion. Previous studies have demonstrated that model capabilities tend to enhance with increased model size, and our findings in the text-to-image generation task align with this conclusion.
%

\subsubsection{Evaluation on Ref Image Creating Tasks}
We evaluate models on face, style, and subject reference creating tasks, to assess their ability for extracting and applying key features from reference images. Performance is measured using four metrics: aesthetic quality (AES), imaging quality (IMG), prompt following (PF) and reference consistency (REF). The results are reported in  \cref{tab:eval_ref_cre}.  

% \noindent\textbf{Face reference generation}. IP-Adapter attains the highest score in reference consistency due to its explicit extraction and integration of facial features during the generation process. Additionally, OmniGen excels in aesthetic quality, image quality, and prompt adherence, while also achieving the second-best performance in reference consistency. These results underscore its superior capability in face reference generation tasks.

% \noindent\textbf{Style reference generation}. All methods exhibit subpar performance. Although ACE and IP-Adapter score highly in reference consistency, their prompt adherence scores are remarkably low, suggesting that these methods fail to comprehend the task, resorting instead to a copy-paste approach. Although OmniGen attains high scores in imaging and aesthetic quality, its performance in reference consistency and prompt adherence is notably low. These findings indicate that current image generation methods possess significant shortcomings in handling style reference generation task.

% \noindent\textbf{Subject reference generation}. We can observe that the prompt following and reference consistency scores showing an obvious trade-off fact: methods that achieve higher reference consistency score generally perform bad on prompt following. Consequently, aesthetic score and imaging quality score are also needed to help judge whether they are good images. 

\subsubsection{Evaluation on No-ref Image Editing Tasks}

Considering the type of source image and the necessity for a source image mask, we further categorize the No-Reference Image Editing tasks into three subgroups: Controllable Generation, Global Editing, and Local Editing. For each subgroup, we compute average scores across evaluation dimensions to determine overall performance.
Aesthetic quality (AES), imaging quality(IMG), and prompt following (PF) are evaluated in all three subgroup. Besides, controllability (CTRL) is assessed in controllable Generation, and source consistency (SRC) is assessed in both global and local editing. 
The experimental results are presented in \cref{tab:eval_noref_edit}.

\subsubsection{Evaluation on Ref Image Editing Tasks}

Like Ref Image Creating tasks, we evaluate the reference consistency using face, style and subject. Notably, we evaluate on virtual try on and subject-guided inpainting tasks for subject reference editing, and average their scores for final performance. Besides, source consistency (SRC) is assessed. The results are provided in \cref{tab:eval_ref_edit}.

\subsection{Model-perspective Evaluation and Comparison}
\label{subsec:model_spec_exp}

\begin{figure*}[t]
  \centering
  \includegraphics[width=\linewidth]{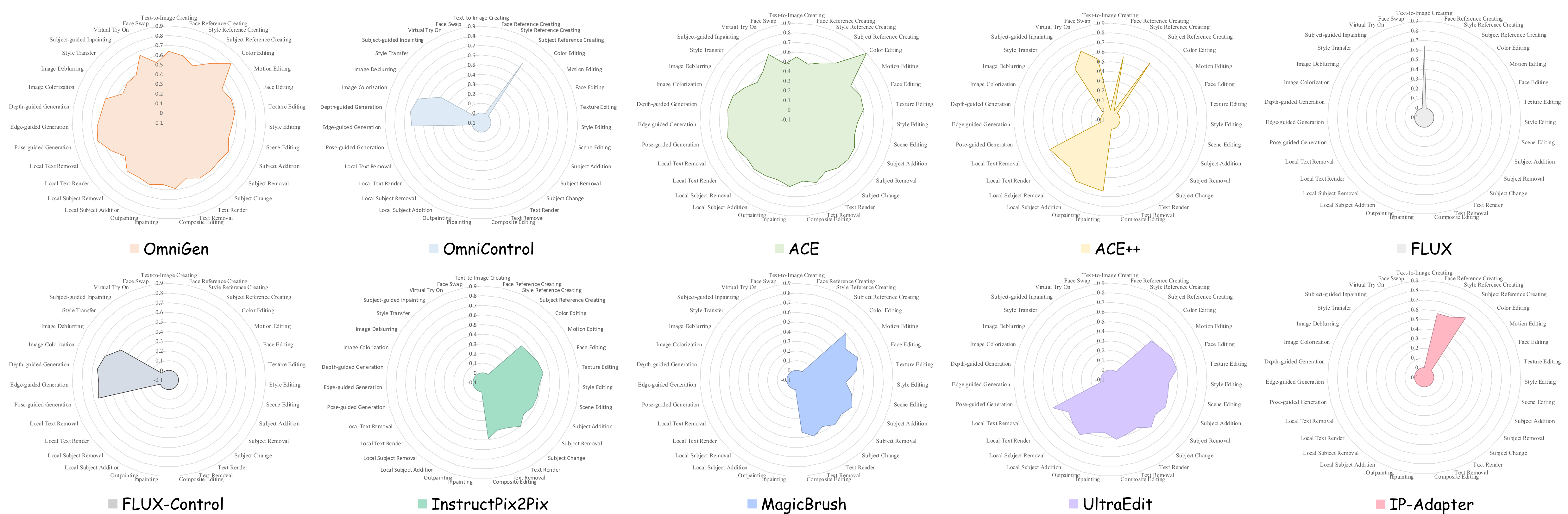}
   \caption{Performance of 10 existing state-of-the-art generation models on 31 evaluation tasks of our \bench.}
   \label{fig:eva}
  \vspace{-10pt}
\end{figure*}

We present a comprehensive evaluation of the models across all 31 tasks. The final score for each task is computed as a weighted sum of scores from multiple evaluation dimensions. Tasks unsupported by a model are assigned a score of zero. The results are visualized in \cref{fig:eva}.

\subsection{Qualitative Analysis of Metric Effectiveness}
%We visualize some test cases together with the generated images from each testing model, to illustrate the effectiveness of the metrics employed in our benchmark. The visualized results are depicted in \cref{fig:case}.

\begin{figure}[h]
  \centering
  \includegraphics[width=\linewidth]{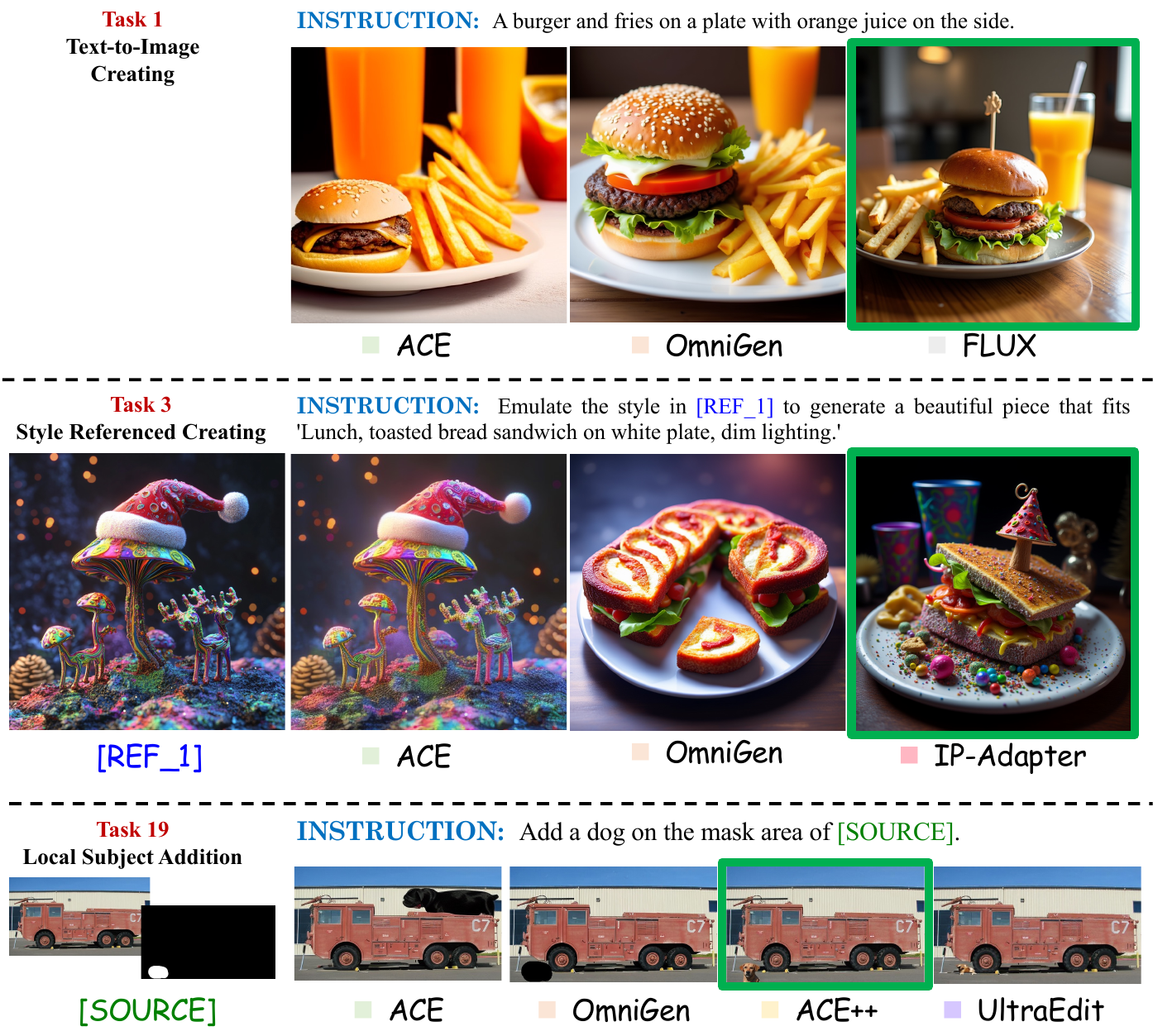}
   \caption{Examples of generation results on 3 tasks.}
   \label{fig:case}
    \vspace{-12pt}
\end{figure}

In \cref{fig:case}, we qualitatively evaluate the effectiveness of our benchmark through three representative cases. In the first case, all models generate images containing a burger, fries, a plate, and orange juice, aligning with the instruction. This aligns with the PF scores in \cref{tab:eval_noref_cre}, where all models achieve comparable results. However, OmniGen and FLUX generate images with superior aesthetic quality and imaging quality, consistent with quantitative evaluations.  
In the style referenced creating, ACE exhibits a “copy-paste" issue, achieving high reference consistency but failing to adhere to the prompt. IP-Adapter strikes a better balance between prompt adherence and reference consistency.  
For local editing, ACE++ outperforms other models, demonstrating precise prompt adherence and high aesthetic quality. In contrast, OmniGen fails to interpret the editing instruction, returning a masked image, which highlights its inadequate training for pixel-aligned editing tasks. These qualitative observations validate the robustness and discriminative power of our metrics.

\section{Insights and Discussions}
\label{sec:dis}

In this section, we present the key observations and insights derived from our comprehensive evaluation experiments.

\noindent\textbf{(1) Limited generality of existing image generation models.}
From \cref{fig:eva}, it is evident that most existing image generation methods are task-specific, with limited generality. Although ACE and OmniGen support all evaluated tasks, their performance remains unsatisfactory in certain areas. Both models achieve low reference consistency scores on the Style Transfer task (0.234 for ACE and 0.359 for OmniGen) and perform poorly on the Face Swap task in terms of reference consistency and aesthetic quality. These limitations underscore the need for developing more versatile and general-purpose generation models capable of handling diverse tasks effectively, and also highlights the importance of conducting unified and comprehensive evaluations, which is exactly what ICE-Bench aims to achieve.

\noindent\textbf{(2) Training data quality and model scalability significantly impact imaging quality.}
Our experiments demonstrate a strong correlation between model scalability, training data quality, and imaging performance. In general, larger models trained with high-resolution images consistently exhibit superior image generation capabilities. For instance, as shown in \cref{tab:eval_noref_cre} and \cref{tab:eval_ref_cre}, OmniGen, with 3.8 billion parameters and trained on images up to $2280\times2280$ resolution, significantly outperforms ACE, which is limited to 0.6 billion parameters and a maximum resolution of $512\times512$.
This disparity underscores the critical role of model capacity and high-quality training data in achieving high-fidelity image generation. These findings suggest that future research should prioritize scaling model architectures and improving data quality over merely increasing data size.

\noindent\textbf{(3) Trade-off across evaluation dimensions.}
As evidenced in \cref{tab:eval_ref_cre}, a notable trade-off exists between reference consistency and prompt adherence. Models that excel in reference consistency, often underperform in prompt adherence. For instance, ACE achieves high reference consistency scores (0.849 for style reference and 0.864 for subject reference) but struggles with prompt adherence, indicating a tendency toward “copy-paste" behavior. This highlights the inherent challenge in balancing precise prompt following with maintaining contextual and stylistic consistency, a critical area for future model improvements.

\noindent\textbf{(4) Effectiveness of pixel-aligned image editing data.}
Regarding SRC, \cref{tab:eval_noref_edit} and \cref{tab:eval_ref_edit} reveal that OmniGen performs significantly worse compared to other models. We attribute this to OmniGen's inadequate training for pixel-aligned image editing tasks. In contrast, ACE, UltraEdit, and ACE++ allocate a higher proportion of training samples to pixel-aligned editing, which enhances their source consistency capabilities. Notably, although InstructPix2Pix and MagicBrush are designed for pixel-aligned editing, their performance is hindered by limited and low-quality training data, resulting in subpar performance across all metrics.

%\noindent\textbf{(5) Qualitative analysis of the metric effectiveness.} 
%In \cref{fig:eva}, we illustrate three cases to qualitatively assess the effectiveness of our benchmark. For the first case, all generated images include elements such as a burger, fries, a plate, and orange juice, accurately aligning with the prompt description. This corresponds to the prompt adherence scores reported in \cref{tab:eval_noref_cre}, where all three methods receive similar scores. However, it is evident that OmniGen and FLUX outperform ACE in terms of aesthetic appeal and imaging quality, which is consistent with the evaluation results.
%In the style reference creation case, ACE encounters a "copy-paste" issue, resulting in high reference consistency but failing to adhere to the given instruction. IP-Adapter performs slightly better, balancing prompt adherence and reference consistency.
%For Local Editing, ACE++ achieves the best performance among the four methods, demonstrating precise prompt adherence and high aesthetic quality. Conversely, OmniGen fails to comprehend the instruction requirements and returns a masked image as output, highlighting its inadequate training for pixel-aligned editing tasks.
\section{Conclusion and Future Work}
\label{sec:con}

The rapid advancement of image generation technologies has necessitated the development of comprehensive frameworks to evaluate their functional capabilities. In this paper, we proposes \textit{a unified and comprehensive benchmark, termed \textbf{\bench}}, to evaluate the generation capabilities of existing models. \bench incorporates \textit{31 fine-grained generation tasks} organized in a coarse-to-fine hierarchy, \textit{6 evaluation dimensions supported by 11 metrics}, and \textit{6,538 task instances} encompassing both real-world and synthetically generated images. This benchmark enables a thorough evaluation of model performance across multiple dimensions, including \textit{aesthetic quality, imaging quality, prompt following, source consistency, reference consistency, and controllability}, thereby fostering innovation and breakthroughs in the field of image generation.

Building on \bench, we are developing a \textit{leaderboard}, a quantifiable and comparative framework that benchmarks the capabilities of existing models, thereby fostering transparency, competition, and continuous advancement in image generation.
\newpage
\small
\bibliographystyle{ieeenat_fullname}
\bibliography{main}

\clearpage
\appendix
\setcounter{figure}{0}
\setcounter{table}{0}
\setcounter{equation}{0}

\section{Details on Evaluation Data}

Our benchmark comprises a total of 31 tasks, with each task containing between 50 and 500 evaluation cases. We provide visualizations of the conditions and exemplary generation results for each task in \cref{fig:evatask}. Specifically, an evaluation case should comprise \textit{(Introduction, Target Caption, Source Image, Source Mask, Reference Images)} to facilitate the generation and evaluation process. A detailed illustration of a complete evaluation case is presented in \cref{tab:supp_case}. 
Most of the existing image generation models support only one or a few of the 31 evaluation tasks. We provide a detailed summary of the tasks supported and unsupported by the 10 evaluated models in \cref{tab:evatask}.

\begin{table}[htb]
\small
\caption{\textbf{Detail of a complete evaluation case.}}
\vspace{-8pt}
\begin{mybox}
    \begin{tabbing}
        \hspace{35mm} \= \kill % Set a tab stop
        \textbf{\texttt{<ItemID>}}: \> b9de809c702c8cf23428ec1 \\
        \> 75af3b0b9 \\
        \textbf{\texttt{<TaskLevel1>}}: \> Reference Editing \\
        \textbf{\texttt{<TaskLevel2>}}: \> Subject Reference Editing \\
        \textbf{\texttt{<Task>}}: \> Subject-guided Inpainting \\
        \textbf{\texttt{<SourceImageType>}}: \> Real Image \\
        \textbf{\texttt{<RegionBased>}}: \> True \\
        \\
        \textbf{\texttt{<SourceImage>}}: \> images/reference\_editing/ \\
        \> subject\_reference\_editing/ \\
        \> subject\_guided\_inpainting/ \\
        \> b9de809c702c8cf234 \\
        \> 28ec175af3b0b9\_src.png \\
        \\
        \textbf{\texttt{<SourceMask>}}: \> images/reference\_editing/ \\
        \> subject\_reference\_editing/ \\
        \> subject\_guided\_inpainting/ \\
        \> b9de809c702c8cf234\\
        \> 28ec175af3b0b9\_mask.png \\
        \\
        \textbf{\texttt{<ReferenceImages>}}: \> ["images/reference\_editing/ \\
        \> subject\_reference\_editing/ \\
        \> subject\_guided\_inpainting/ \\
        \> b9de809c702c8cf234 \\
        \> 28ec175af3b0b9\_ref1.png"] \\
        \\
        \textbf{\texttt{<Instruction>}}: \> Take \texttt{<REF\_1>} as a \\
        \> reference to repaint the \\
        \> masked part of \texttt{<SOURCE>}. \\
        \\
        \textbf{\texttt{<SourceCaption>}}: \> Eye-level view of a street \\
        \>  scene featuring a fire \\
        \>  hydrant in the foreground. \\
        \\
        \textbf{\texttt{<TargetCaption>}}: \> A small, brightly colored toy \\
        \> car sits on a weathered asphalt \\
        \> surface, positioned slightly \\
        \> off-center in the foreground. \\
        \> The car is predominantly red \\
        \> and yellow, with green accents. \\
    \end{tabbing}
\end{mybox}
\label{tab:supp_case}
\end{table}
\begin{table*}[!t]
  \centering
  \caption{Task-model correspondence.}
  \resizebox{\textwidth}{!}{
  \footnotesize
  \setlength{\tabcolsep}{0.5pt}
      \begin{tabular}{@{}c|c|c|c|cccccccccc@{}}
        \toprule
        \multicolumn{4}{c|}{Evaluation Tasks} & OmniGen~\cite{OmniGen} & ACE~\cite{ACE} & FLUX~\cite{FLUX} & OminiControl~\cite{OminiControl} & InstructPix2Pix~\cite{IP2P} & MagicBrush~\cite{MagicBrush} & UltraEdit~\cite{UltraEdit} & FLUX-Control~\cite{FLUX-Control} & IP-Adapter~\cite{IP-Adapter} & ACE++~\cite{ACE++}\\
        \midrule
        \multirow{4}{*}{Creating} & No-Ref & \multicolumn{2}{c|}{(1) Text-to-Image Creating} & \color{green}{\checkmark} & \color{green}{\checkmark} &\color{green}{\checkmark} &\color{red}{\texttimes}&\color{red}{\texttimes}&\color{red}{\texttimes}&\color{red}{\texttimes}&\color{red}{\texttimes}&\color{red}{\texttimes}&\color{red}{\texttimes}\\
        \cline{2-4}
        & \multirow{3}{*}{Ref} & \multicolumn{2}{c|}{(2) Face Reference Creating} & \color{green}{\checkmark} & \color{green}{\checkmark} &\color{red}{\texttimes}&\color{red}{\texttimes}&\color{red}{\texttimes}&\color{red}{\texttimes}&\color{red}{\texttimes}&\color{red}{\texttimes}& \color{green}{\checkmark} & \color{green}{\checkmark} \\
        &  & \multicolumn{2}{c|}{(3) Style Reference Creating}& \color{green}{\checkmark} & \color{green}{\checkmark} &\color{red}{\texttimes}&\color{red}{\texttimes}&\color{red}{\texttimes}&\color{red}{\texttimes}&\color{red}{\texttimes}&\color{red}{\texttimes}&\color{green}{\checkmark} &\color{red}{\texttimes}\\
        &  & \multicolumn{2}{c|}{(4) Subject Reference Creating}& \color{green}{\checkmark} & \color{green}{\checkmark} &\color{red}{\texttimes}& \color{green}{\checkmark} &\color{red}{\texttimes}&\color{red}{\texttimes}&\color{red}{\texttimes}&\color{red}{\texttimes}&\color{green}{\checkmark} & \color{green}{\checkmark} \\
        \cline{1-14}
        \multirow{27}{*}{Editing} & \multirow{23}{*}{No-Ref} &  \multirow{12}{*}{Global} & (5) Color Editing& \color{green}{\checkmark}& \color{green}{\checkmark}&\color{red}{\texttimes}&\color{red}{\texttimes}&\color{green}{\checkmark}&\color{green}{\checkmark}&\color{green}{\checkmark}&\color{red}{\texttimes}&\color{red}{\texttimes}&\color{red}{\texttimes}\\
        &&&(6) Motion Editing & \color{green}{\checkmark}& \color{green}{\checkmark}&\color{red}{\texttimes}&\color{red}{\texttimes}&\color{green}{\checkmark}&\color{green}{\checkmark}&\color{green}{\checkmark}&\color{red}{\texttimes}&\color{red}{\texttimes}&\color{red}{\texttimes}\\
        &&&(7) Face Editing & \color{green}{\checkmark}& \color{green}{\checkmark}&\color{red}{\texttimes}&\color{red}{\texttimes}&\color{green}{\checkmark}&\color{green}{\checkmark}&\color{green}{\checkmark}&\color{red}{\texttimes}&\color{red}{\texttimes}&\color{red}{\texttimes}\\
        &&&(8) Texture Editing & \color{green}{\checkmark}& \color{green}{\checkmark}&\color{red}{\texttimes}&\color{red}{\texttimes}&\color{green}{\checkmark}&\color{green}{\checkmark}&\color{green}{\checkmark}&\color{red}{\texttimes}&\color{red}{\texttimes}&\color{red}{\texttimes}\\
        &&&(9) Style Editing & \color{green}{\checkmark}& \color{green}{\checkmark}&\color{red}{\texttimes}&\color{red}{\texttimes}&\color{green}{\checkmark}&\color{green}{\checkmark}&\color{green}{\checkmark}&\color{red}{\texttimes}&\color{red}{\texttimes}&\color{red}{\texttimes}\\
        &&&(10) Scene Editing & \color{green}{\checkmark}& \color{green}{\checkmark}&\color{red}{\texttimes}&\color{red}{\texttimes}&\color{green}{\checkmark}&\color{green}{\checkmark}&\color{green}{\checkmark}&\color{red}{\texttimes}&\color{red}{\texttimes}&\color{red}{\texttimes}\\
        &&&(11) Subject Addition & \color{green}{\checkmark}& \color{green}{\checkmark}&\color{red}{\texttimes}&\color{red}{\texttimes}&\color{green}{\checkmark}&\color{green}{\checkmark}&\color{green}{\checkmark}&\color{red}{\texttimes}&\color{red}{\texttimes}&\color{red}{\texttimes}\\
        &&&(12) Subject Removal & \color{green}{\checkmark}& \color{green}{\checkmark}&\color{red}{\texttimes}&\color{red}{\texttimes}&\color{green}{\checkmark}&\color{green}{\checkmark}&\color{green}{\checkmark}&\color{red}{\texttimes}&\color{red}{\texttimes}&\color{red}{\texttimes}\\
        &&&(13) Subject Change & \color{green}{\checkmark}& \color{green}{\checkmark}&\color{red}{\texttimes}&\color{red}{\texttimes}&\color{green}{\checkmark}&\color{green}{\checkmark}&\color{green}{\checkmark}&\color{red}{\texttimes}&\color{red}{\texttimes}&\color{red}{\texttimes}\\
        &&&(14) Text Render & \color{green}{\checkmark}& \color{green}{\checkmark}&\color{red}{\texttimes}&\color{red}{\texttimes}&\color{green}{\checkmark}&\color{green}{\checkmark}&\color{green}{\checkmark}&\color{red}{\texttimes}&\color{red}{\texttimes}&\color{red}{\texttimes}\\
        &&&(15) Text Removal & \color{green}{\checkmark}& \color{green}{\checkmark}&\color{red}{\texttimes}&\color{red}{\texttimes}&\color{green}{\checkmark}&\color{green}{\checkmark}&\color{green}{\checkmark}&\color{red}{\texttimes}&\color{red}{\texttimes}&\color{red}{\texttimes}\\
        &&&(16) Composite Editing & \color{green}{\checkmark}& \color{green}{\checkmark}&\color{red}{\texttimes}&\color{red}{\texttimes}&\color{green}{\checkmark}&\color{green}{\checkmark}&\color{green}{\checkmark}&\color{red}{\texttimes}&\color{red}{\texttimes}&\color{red}{\texttimes}\\
        \cline{3-4}
        &  & \multirow{6}{*}{Local} & (17) Inpainting& \color{green}{\checkmark}& \color{green}{\checkmark}&\color{red}{\texttimes}&\color{red}{\texttimes}&\color{red}{\texttimes}&\color{red}{\texttimes}&\color{green}{\checkmark}&\color{red}{\texttimes}&\color{red}{\texttimes}& \color{green}{\checkmark}\\
        &&&(18) Outpainting & \color{green}{\checkmark}& \color{green}{\checkmark}&\color{red}{\texttimes}&\color{red}{\texttimes}&\color{red}{\texttimes}&\color{red}{\texttimes}&\color{green}{\checkmark}&\color{red}{\texttimes}&\color{red}{\texttimes}& \color{green}{\checkmark}\\
        &&&(19) Local Subject Addition & \color{green}{\checkmark}& \color{green}{\checkmark}&\color{red}{\texttimes}&\color{red}{\texttimes}&\color{red}{\texttimes}&\color{red}{\texttimes}&\color{green}{\checkmark}&\color{red}{\texttimes}&\color{red}{\texttimes}& \color{green}{\checkmark}\\
        &&&(20) Local Subject Removal & \color{green}{\checkmark}& \color{green}{\checkmark}&\color{red}{\texttimes}&\color{red}{\texttimes}&\color{red}{\texttimes}&\color{red}{\texttimes}&\color{green}{\checkmark}&\color{red}{\texttimes}&\color{red}{\texttimes}& \color{green}{\checkmark}\\
        &&&(21) Local Text Removal & \color{green}{\checkmark}& \color{green}{\checkmark}&\color{red}{\texttimes}&\color{red}{\texttimes}&\color{red}{\texttimes}&\color{red}{\texttimes}&\color{green}{\checkmark}&\color{red}{\texttimes}&\color{red}{\texttimes}& \color{green}{\checkmark}\\
        &&&(22) Local Text Render & \color{green}{\checkmark}& \color{green}{\checkmark}&\color{red}{\texttimes}&\color{red}{\texttimes}&\color{red}{\texttimes}&\color{red}{\texttimes}&\color{green}{\checkmark}&\color{red}{\texttimes}&\color{red}{\texttimes}& \color{green}{\checkmark}\\
        \cline{2-4}
        &  & \multirow{5}{*}{Controllable} & (23) Pose-guided Generation &\color{green}{\checkmark} &\color{green}{\checkmark} &\color{red}{\texttimes}&\color{red}{\texttimes}&\color{red}{\texttimes}&\color{red}{\texttimes}&\color{red}{\texttimes}&\color{green}{\checkmark}&\color{red}{\texttimes}&\color{red}{\texttimes}\\
        &&&(24) Edge-guided Generation &\color{green}{\checkmark} &\color{green}{\checkmark} &\color{red}{\texttimes}&\color{green}{\checkmark}&\color{red}{\texttimes}&\color{red}{\texttimes}&\color{red}{\texttimes}&\color{green}{\checkmark}&\color{red}{\texttimes}&\color{red}{\texttimes}\\
        &&&(25) Depth-guided Generation &\color{green}{\checkmark} &\color{green}{\checkmark} &\color{red}{\texttimes}&\color{green}{\checkmark}&\color{red}{\texttimes}&\color{red}{\texttimes}&\color{red}{\texttimes}&\color{green}{\checkmark}&\color{red}{\texttimes}&\color{red}{\texttimes}\\
        &&&(26) Image Colorization &\color{green}{\checkmark} &\color{green}{\checkmark} &\color{red}{\texttimes}&\color{green}{\checkmark}&\color{red}{\texttimes}&\color{red}{\texttimes}&\color{red}{\texttimes}&\color{green}{\checkmark}&\color{red}{\texttimes}&\color{red}{\texttimes}\\
        &&&(27) Image Deblurring &\color{green}{\checkmark} &\color{green}{\checkmark} &\color{red}{\texttimes}&\color{green}{\checkmark}&\color{red}{\texttimes}&\color{red}{\texttimes}&\color{red}{\texttimes}&\color{green}{\checkmark}&\color{red}{\texttimes}&\color{red}{\texttimes}\\
        \cline{3-4}  
        & \multirow{4}{*}{Ref} & \multicolumn{2}{c|}{(28) Style Transfer} & \color{green}{\checkmark}& \color{green}{\checkmark}&\color{red}{\texttimes}&\color{red}{\texttimes}&\color{red}{\texttimes}&\color{red}{\texttimes}&\color{red}{\texttimes}&\color{red}{\texttimes}&\color{red}{\texttimes}&\color{red}{\texttimes}\\
        \cline{3-4}
        & & \multirow{2}{*}{Subject} & (29) Subject-guided Inpainting & \color{green}{\checkmark}& \color{green}{\checkmark}&\color{red}{\texttimes}&\color{red}{\texttimes}&\color{red}{\texttimes}&\color{red}{\texttimes}&\color{red}{\texttimes}&\color{red}{\texttimes}&\color{red}{\texttimes}&\color{green}{\checkmark}\\
        & & & (30) Virtual Try On & \color{green}{\checkmark}& \color{green}{\checkmark}&\color{red}{\texttimes}&\color{red}{\texttimes}&\color{red}{\texttimes}&\color{red}{\texttimes}&\color{red}{\texttimes}&\color{red}{\texttimes}&\color{red}{\texttimes}&\color{green}{\checkmark}\\
        \cline{3-4}
        & & \multicolumn{2}{c|}{(31) Face Swap} & \color{green}{\checkmark}& \color{green}{\checkmark}&\color{red}{\texttimes}&\color{red}{\texttimes}&\color{red}{\texttimes}&\color{red}{\texttimes}&\color{red}{\texttimes}&\color{red}{\texttimes}&\color{red}{\texttimes}&\color{green}{\checkmark} \\
        \bottomrule
      \end{tabular}
  }
  \label{tab:evatask}
\end{table*}

\section{Details on Evaluation Dimensions}

\subsection{Aesthetic Quality}

\begin{figure}[!h]
  \centering
  \includegraphics[width=\linewidth]{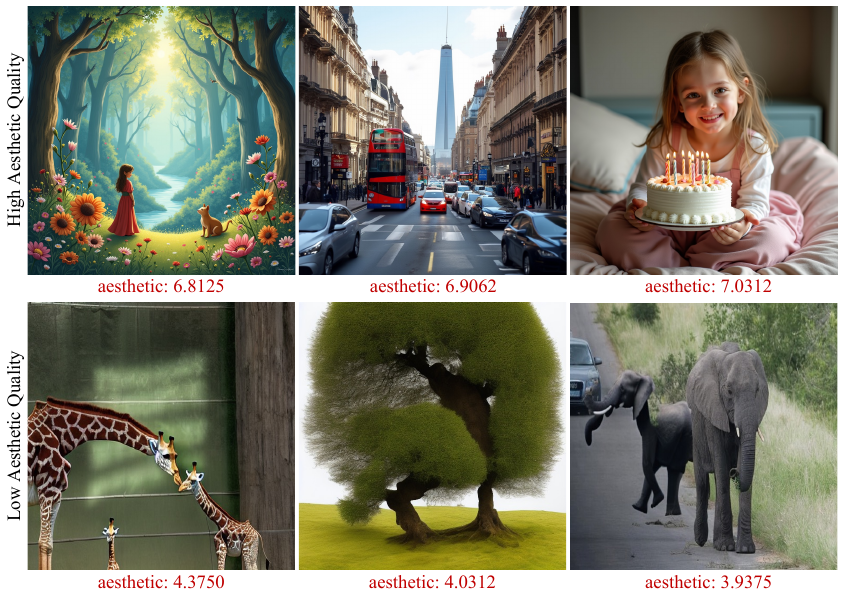}
   \caption{\textbf{Visualization of Aesthetic Quality.} Images that receive high aesthetic scores exhibit artistic appeal, whereas those with low aesthetic scores tend to appear unattractive.}
   \label{fig:supp_aes_case}
\end{figure}

Aesthetic Quality evaluates the principles of photographic composition, considering color harmony, subject arrangement, and the overall artistic impression of the image. We utilize a SigLip-based image aesthetic quality predictor to assess the aesthetic score of the generated image. The model produces a rating on a scale from 0 to 10, which we linearly normalize to a range of [0, 1] by dividing the raw score by 10.

\begin{equation}
    S_{\text{AES}} = \frac{f_{\text{AES}}(\mathbf{I})}{10}
\end{equation}

\subsection{Imaging Quality}

\begin{figure}[ht]
  \centering
  \includegraphics[width=\linewidth]{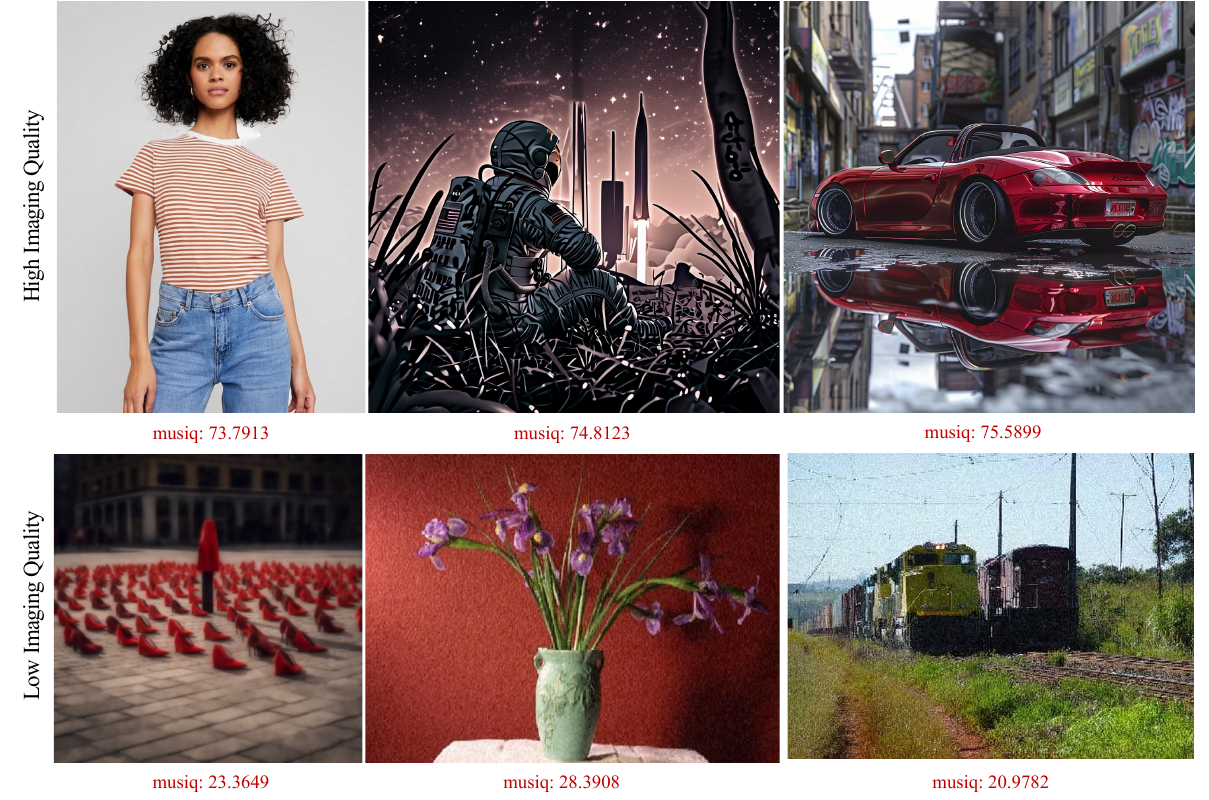}
   \caption{\textbf{Visualization of Imaging Quality.} Images that achieve high imaging quality scores are typically clear and possess sharp edges, whereas those with low scores tend to appear blurry and noisy.}
   \label{fig:supp_img_case}
\end{figure}

Imaging quality primarily examines the low-level characteristics of the generated image, such as edge sharpness, distortion, overexposure, noise, and blur. We employ the MUSIQ image quality predictor trained on the Koniq dataset, as implemented in IQA-Pytorch~\cite{IQA-Pytorch}. For consistency and fairness in comparison, we resize the height of all generated images to 1024 pixels before inputting them into the model to assess imaging quality. This approach inherently favors high-resolution images as they typically exhibit superior imaging quality compared to low-resolution images.
The model produces a score on a scale from 0 to 100, which we linearly normalize to a range of [0, 1] by dividing the raw score by 100.

\begin{equation}
    S_{\text{IMG}} = \frac{f_{\text{MUSIQ}}(\mathbf{I})}{100}
\end{equation}

\subsection{Prompt Following}

\begin{figure}[ht]
  \centering
  \includegraphics[width=\linewidth]{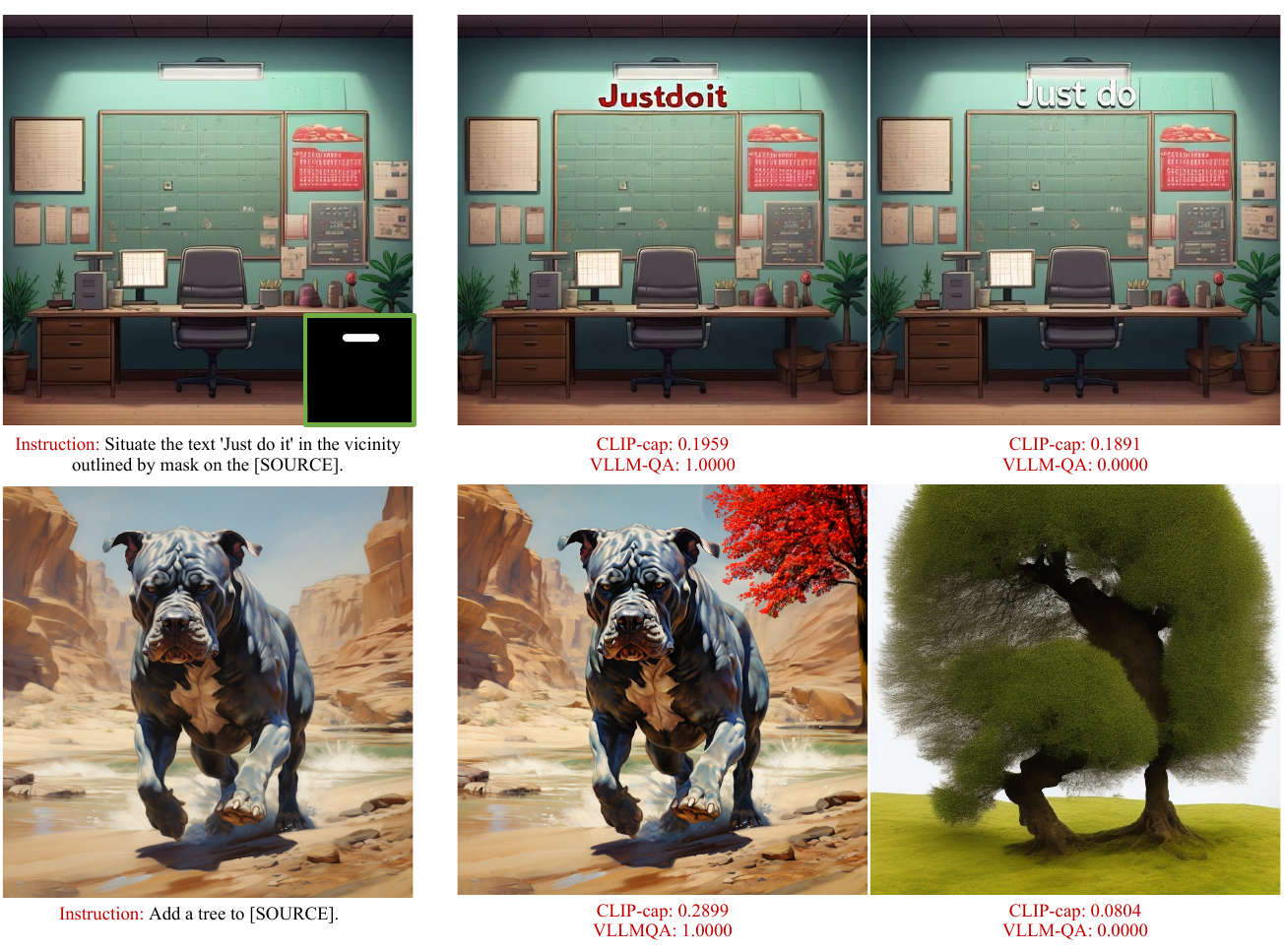}
   \caption{\textbf{Visualization of Prompt Following.} Both the CLIP-cap and VLLM-QA metrics effectively capture the successful execution of instructions.}
   \label{fig:supp_pf_case}
\end{figure}

The prompt-following score evaluates the degree to which the generated image aligns with the provided textual instructions or descriptions. For image creation tasks and controllable generation tasks, we compute the CLIP~\cite{CLIP} similarity between the target caption and the generated image directly. The prompt-following score is then obtained by normalizing the CLIP similarity, specifically by dividing it by 0.5.

\begin{equation}
    S_{\text{PF}} = \frac{\langle d_{\text{prompt}} \cdot d_{\mathbf{I}} \rangle}{0.5}
\end{equation}

Notably, for the Image Colorization and Image Deblurring tasks, CLIP similarity alone is insufficient to accurately assess prompt-following capability. For the Image Colorization task, the colorfulness score must also be considered an essential metric, leading us to adapt the prompt-following score accordingly:

\begin{equation}
    S_{\text{PF}}^{\text{colorsize}} = \frac{\langle d_{\text{prompt}} \cdot d_{\mathbf{I}} \rangle}{0.5} + s_{\text{color}}
\end{equation}

In the case of the Image Deblurring task, the Imaging score serves as the prompt-following metric, as the primary objective is to enhance image quality.

\begin{equation}
    S_{\text{PF}}^{\text{deblur}} = S_{\text{IMG}}
\end{equation}

For image editing tasks, relying solely on CLIP similarity is insufficient to determine whether instructions have been correctly executed. To address this, we introduce a novel VLLM-based metric called VLLM-QA to assess the success of instruction alignment. We employ the QWEN2-VL-72B~\cite{QWEN2VL} model as our QA tool, prompting it with all relevant input components, including the instruction, source image, reference images, source mask, and the generated image. The model is tasked with evaluating whether the instruction has been accurately implemented; it returns a score of 1 for success and 0 otherwise. We calculate the VLLM-QA score by averaging the results across all cases within a task. Subsequently, the prompt-following score is determined as follows:

\begin{equation}
    S_{\text{PF}} = \frac{\frac{\langle d_{\text{prompt}} \cdot d_{\mathbf{I}} \rangle}{0.5} + f_{\text{QWEN}}(\cdot)}{2}
\end{equation}

\subsection{Source Consistency}

\begin{figure}[ht]
  \centering
  \includegraphics[width=\linewidth]{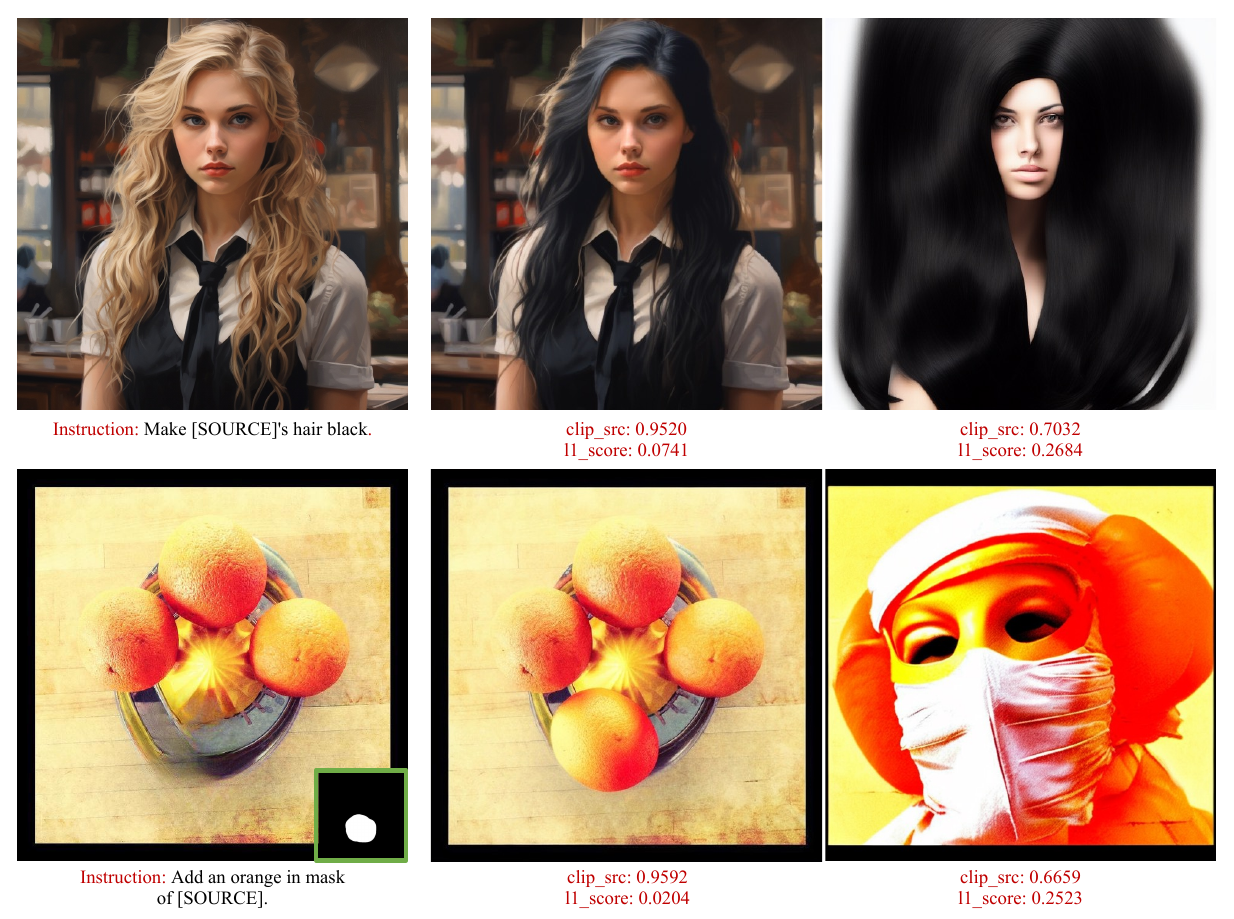}
   \caption{\textbf{Visualization of Source Consistency.} Images that exhibit strong pixel alignment with the source image attain higher CLIP-src scores and lower L1 scores. These outcomes underscore the effectiveness of our evaluation of Source Consistency.}
   \label{fig:supp_src_case}
\end{figure}

For image editing tasks, it is crucial to maintain the pixels that are unrelated to the editing instructions unchanged. To evaluate the models' ability to preserve pixel alignment, we compute both the CLIP similarity and the mean L1 distance between the generated image and the source image. The Source Consistency score is then calculated as follows:

\begin{equation}
    S_{\text{SRC}} = \frac{\langle d_{\mathbf{I}_{\text{src}}} \cdot d_{\mathbf{I}} \rangle + 1 - L1(\mathbf{I}_{\text{src}}, \mathbf{I})}{2}
\end{equation}

\subsection{Reference Consistency}

\begin{figure}[ht]
  \centering
  \includegraphics[width=\linewidth]{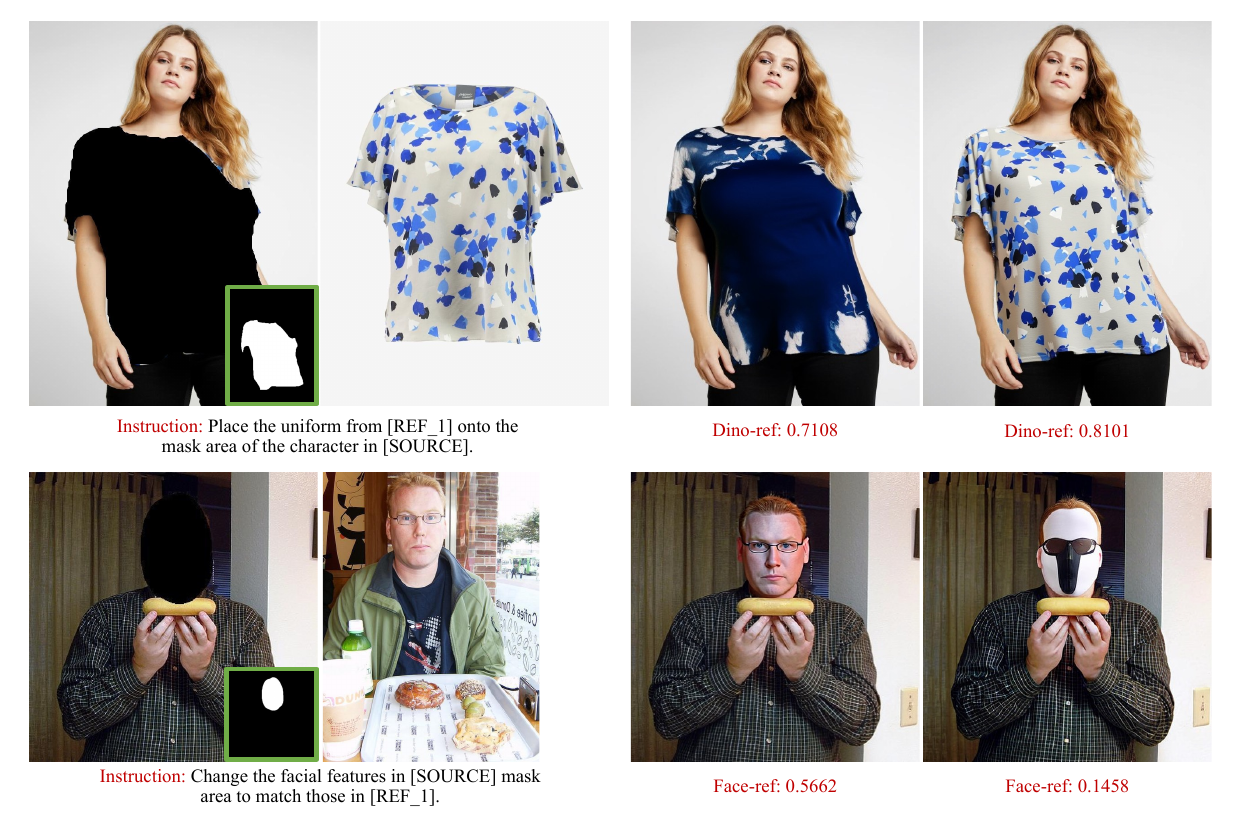}
   \caption{\textbf{Visualization of Reference Consistency.} Images that maintain identity preservation with the reference image achieve higher CLIP-ref scores, highlighting the effectiveness of our Reference Consistency evaluation.}
   \label{fig:supp_ref_case}
\end{figure}

Reference consistency evaluates the semantic alignment between the reference image and the generated image across specific aspects, such as face, style, and subject. To achieve this, we utilize different encoders to extract embeddings from both the reference image and the generated image. We then assess the reference consistency in these three dimensions by calculating the feature similarity between the extracted embeddings:

\begin{equation}
    S_{\text{REF}} = \langle d_{\mathbf{I}_{\text{ref}}} \cdot d_{\mathbf{I}} \rangle
\end{equation}

\subsection{Controllability}

\begin{figure}[ht]
  \centering
  \includegraphics[width=\linewidth]{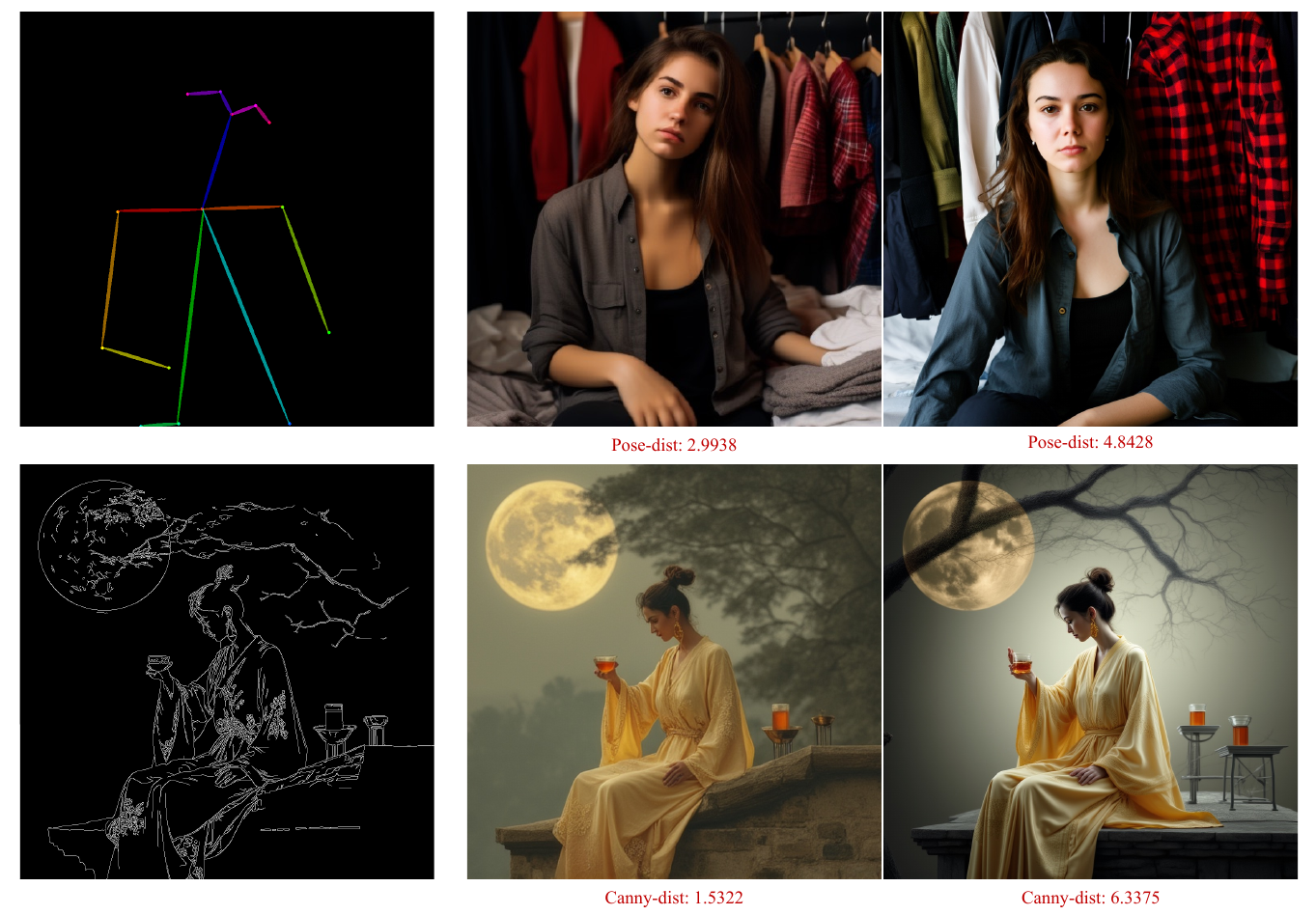}
   \caption{\textbf{Visualization of Controllability.} The Pose-dist and Canny-dist metrics effectively indicate controllability, with lower values generally signifying greater controllability.}
   \label{fig:supp_ctrl_case}
\end{figure}

Controllability evaluates the alignment of low-level features in the generated image with the input condition image. For tasks such as Pose, Depth, Edge-guided Generation, and Image Colorization, we extract the relevant low-level feature map from the generated image and calculate the mean L1 score between this feature map and the input condition image. The controllability score is then determined as follows:

\begin{equation}
    S_{\text{CTRL}} = 1 - (f_{\text{enc}}(\mathbf{I}) - \mathbf{I}_{\text{src}})
\end{equation}

While for Image Deblurring task, we employ the SSIM score as the controllability score:

\begin{equation}
    S_{\text{CTRL}}^{\text{deblur}} = \text{SSIM}(\mathbf{I}, \mathbf{I}_{\text{src}})
\end{equation}

\section{Details on Model Performance per Task}
In this section, we present the detailed evaluation results for each metric across all tasks and models. The results for No-ref Image Creating are shown in \cref{tab:supp_eval_noref_cre_metric}. The results for Ref Image Creating are provided in \cref{tab:supp_eval_ref_cre_metric}. For No-ref Image Editing, the results are detailed in \cref{tab:supp_eval_noref_edit_metric_controllable}, \cref{tab:supp_eval_noref_edit_metric_global}, and \cref{tab:supp_eval_noref_edit_metric_local}. The results for Ref Image Editing are reported in \cref{tab:supp_eval_ref_edit_metric}.

\begin{table}[ht]
  \centering
  \caption{Metrics on No-ref Image Creating Task (Task 1).}
  \vspace{-8pt}
  \small
  \setlength{\tabcolsep}{4pt}
  \begin{tabular}{c|ccc}
    \toprule
    Models & Aesthetic Score$\uparrow$ & Imaging Score$\uparrow$ & CLIP-cap$\uparrow$ \\
    \midrule
    ACE & 5.485  & 53.403  &  0.283 \\
    OmniGen & 6.107  & 72.615  & \textbf{0.285}  \\
    FLUX & \textbf{6.175}  & \textbf{73.480}  & \textbf{0.285}  \\
    \bottomrule
  \end{tabular}
  \label{tab:supp_eval_noref_cre_metric}
\end{table}
\begin{table}[ht]
  \centering
  \caption{Metrics on Controllable Generation Tasks (Tasks 23-27).}
  \vspace{-8pt}
  \small
  \begin{minipage}{\columnwidth}
     %subtable1
    \setlength{\tabcolsep}{4pt}{
      \begin{tabular}{p{2cm}|cccc}
        \toprule
        \multirow{2}{*}{Models} & \multicolumn{4}{c}{\textbf{Task 23: Pose-guided Generation}}  \\
        & \makecell{Aesthetic\\Score}$\uparrow$ & \makecell{Imaging\\Score}$\uparrow$ & CLIP-cap$\uparrow$ & L1-src$\downarrow$\\
        \midrule 
        ACE & \textbf{5.568} & 50.253 & \textbf{0.299} & \textbf{0.009} \\
        OmniGen & 5.365& \textbf{61.463} & 0.298 & 0.015 \\
        FLUX-Control & 5.538 & 56.010 & 0.298 & 0.015 \\
        \bottomrule
      \end{tabular}
      }

    %subtable2
    \setlength{\tabcolsep}{4pt}{
      \begin{tabular}{p{2cm}|cccc}
        \toprule
        \multirow{2}{*}{Models} & \multicolumn{4}{c}{\textbf{Task 24: Edge-guided Generation}}  \\
        & \makecell{Aesthetic\\Score}$\uparrow$ & \makecell{Imaging\\Score}$\uparrow$ & CLIP-cap$\uparrow$ & L1-src$\downarrow$\\
        \midrule 
        ACE & 5.319 & 49.506 & 0.298 & 0.091  \\
        OmniGen & 4.897 & \textbf{66.168} & 0.293 & 0.102  \\
        FLUX-Control & 5.493 & 54.225 & 0.296 & 0.104  \\
        OminiControl & \textbf{5.507} & 51.301 & \textbf{0.299} & \textbf{0.087} \\
        \bottomrule
      \end{tabular}
      }

      %subtable3
    \setlength{\tabcolsep}{4pt}{
      \begin{tabular}{p{2cm}|cccc}
        \toprule
        \multirow{2}{*}{Models} & \multicolumn{4}{c}{\textbf{Task 25: Depth-guided Generation}}  \\
        & \makecell{Aesthetic\\Score}$\uparrow$ & \makecell{Imaging\\Score}$\uparrow$ & CLIP-cap$\uparrow$ & L1-src$\downarrow$\\
        \midrule 
        ACE & 5.505 & 51.948 & 0.291 & \textbf{0.095}   \\
        OmniGen & 4.809 & \textbf{60.266} & 0.266 & 0.131 \\
        FLUX-Control & \textbf{5.844} & 59.578 & 0.295 & 0.123   \\
        OminiControl & 5.762 & 57.305 & \textbf{0.296} & 0.098 \\
        \bottomrule
      \end{tabular}
      }

      %subtable4
    \setlength{\tabcolsep}{1.3pt}{
      \begin{tabular}{p{2cm}|ccccc}
        \toprule
        \multirow{2}{*}{Models} & \multicolumn{5}{c}{\textbf{Task 26: Image Colorization}}  \\
        & \makecell{Aesthetic\\Score}$\uparrow$ & \makecell{Imaging\\Score}$\uparrow$ & CLIP-cap$\uparrow$ & \makecell{Color\\Score}$\uparrow$ & L1-src$\downarrow$\\
        \midrule 
        ACE & 5.325 & 50.484 & 0.295 & \textbf{0.278} & 0.059    \\
        OmniGen & 5.275 & \textbf{61.076} & 0.289 & 0.189 & 0.185 \\
        FLUX-Control & \textbf{5.371} & 51.891 & \textbf{0.302} & 0.210 & 0.067 \\
        OminiControl & 5.272 & 50.995 & 0.301 & 0.161 & \textbf{0.029} \\
        \bottomrule
      \end{tabular}
      }

      %subtable5
    \setlength{\tabcolsep}{9.5pt}{
      \begin{tabular}{p{2cm}|ccc}
        \toprule
        \multirow{2}{*}{Models} & \multicolumn{3}{c}{\textbf{Task 27: Image Deblurring}}  \\
        & \makecell{Aesthetic\\Score}$\uparrow$ & \makecell{Imaging\\Score}$\uparrow$ & SSIM$\uparrow$ \\
        \midrule 
        ACE & \textbf{5.556} & \textbf{50.229} & 0.582 \\
        OmniGen & 5.133 & 48.144 & 0.350 \\
        FLUX-Control & 5.342 & 45.063 & 0.540  \\
        OminiControl & 4.249 & 30.327 &	\textbf{0.650} \\
        \bottomrule
      \end{tabular}
      }
  \end{minipage}
  \label{tab:supp_eval_noref_edit_metric_controllable}
\end{table}
\begin{table}[ht]
  \centering
  \caption{Metrics on Ref Image Editing Tasks (Tasks 28-31).}
  \vspace{-8pt}
  \footnotesize
  \setlength{\tabcolsep}{1.5pt}{
     %subtable1
      \begin{tabular}{c|ccccccc}
        \toprule
        \multirow{2}{*}{Models} & \multicolumn{7}{c}{\textbf{Task 28: Style Transfer}}  \\
        & \makecell{Aesthetic\\Score}$\uparrow$ & \makecell{Imaging\\Score}$\uparrow$ & \makecell{CLIP\\-cap}$\uparrow$ &\makecell{VLLM\\-QA}$\uparrow$ &\makecell{Style\\-ref}$\uparrow$ &  \makecell{CLIP\\-src}$\uparrow$ &L1-src$\downarrow$\\
        \midrule 
        ACE & \textbf{5.346} & 53.030 & 0.189 & \textbf{0.323} & 0.234 & \textbf{0.762} & \textbf{0.186} \\
        OmniGen & 5.045 & \textbf{62.995} & \textbf{0.193} & 0.290 & \textbf{0.359} & 0.680 & 0.277   \\
        \bottomrule
      \end{tabular}

    %subtable2
    % \resizebox{\textwidth}{!}{
      \begin{tabular}{c|ccccccc}
        \toprule
        \multirow{2}{*}{Models} & \multicolumn{7}{c}{\textbf{Task 29: Subject-guided Inpainting}}  \\
        & \makecell{Aesthetic\\Score}$\uparrow$ & \makecell{Imaging\\Score}$\uparrow$ & \makecell{CLIP\\-cap}$\uparrow$ &\makecell{VLLM\\-QA}$\uparrow$ &\makecell{DINO\\-ref}$\uparrow$ &  \makecell{CLIP\\-src}$\uparrow$ &L1-src$\downarrow$\\
        \midrule 
        ACE & 4.812 & 52.544 & \textbf{0.197} & 0.171 & 0.562 & \textbf{0.766} & \textbf{0.015} \\
        OmniGen & 4.459 & 59.995 & 0.186 & 0.093 & 0.555 & 0.642 & 0.149 \\
        ACE++ & \textbf{4.835} & \textbf{63.419} & 0.186 & \textbf{0.257} & \textbf{0.563} & 0.753 & 0.040 \\
        \bottomrule
      \end{tabular}
      % }

      %subtable4
    % \resizebox{\textwidth}{!}{
      \begin{tabular}{c|ccccccc}
        \toprule
        \multirow{2}{*}{Models} & \multicolumn{7}{c}{\textbf{Task 31: Face Swap}}  \\
        & \makecell{Aesthetic\\Score}$\uparrow$ & \makecell{Imaging\\Score}$\uparrow$ & \makecell{CLIP\\-cap}$\uparrow$ &\makecell{VLLM\\-QA}$\uparrow$ &\makecell{Face\\-ref}$\uparrow$ &  \makecell{CLIP\\-src}$\uparrow$ &L1-src$\downarrow$\\
        \midrule 
        ACE & 4.983 & 56.985 & \textbf{0.232} & 0.400 & 0.250 & \textbf{0.763} & \textbf{0.018} \\
        OmniGen & 4.309 & 64.021 & 0.217 & \textbf{0.484} & \textbf{0.477} & 0.661 & 0.112   \\
        ACE++ & \textbf{5.034} & \textbf{64.963} & 0.231 & 0.442 & 0.378 & 0.760 & 0.054  \\
        \bottomrule
      \end{tabular}
      % }

      %subtable3
    % \resizebox{\textwidth}{!}{
      \begin{tabular}{c|ccccccc}
        \toprule
        \multirow{2}{*}{Models} & \multicolumn{7}{c}{\textbf{Task 30: Virtual Try On}}  \\
        & \makecell{Aesthetic\\Score}$\uparrow$ & \makecell{Imaging\\Score}$\uparrow$ & \makecell{CLIP\\-cap}$\uparrow$ &\makecell{VLLM\\-QA}$\uparrow$ &\makecell{DINO\\-ref}$\uparrow$ &  \makecell{CLIP\\-src}$\uparrow$ &L1-src$\downarrow$\\
        \midrule 
        ACE & \textbf{4.837} & 64.723 & 0.231 & 0.629 & 0.751 & \textbf{0.889} & \textbf{0.006} \\
        OmniGen & 4.696 & 73.313 & 0.235 & 0.722 & 0.744 & 0.847 & 0.058  \\
        ACE++ & 4.577 & \textbf{73.525} & \textbf{0.243} & \textbf{0.804} & \textbf{0.763} & 0.882 & 0.029 \\
        \bottomrule
      \end{tabular}
      % }
  }
  \label{tab:supp_eval_ref_edit_metric}
\end{table}

\begin{table*}[ht]
  \centering
  \caption{Metrics on Ref Image Creating Tasks (Tasks 2-4).}
  \vspace{-8pt}
  \footnotesize
  \setlength{\tabcolsep}{1.3pt}{
      \begin{tabular}{c|cccc|cccc|cccc}
        \toprule
        \multirow{2}{*}{Models} & \multicolumn{4}{c}{\textbf{Task 2: Face Reference Creating}} & \multicolumn{4}{c}{\textbf{Task 3: Style Reference Creating}} & \multicolumn{4}{c}{\textbf{Task 4: Subject Reference Creating}} \\
        & \makecell{Aesthetic \\Score}$\uparrow$ & \makecell{Imaging\\ Score}$\uparrow$ & CLIP-cap$\uparrow$ & Face-ref$\uparrow$ & \makecell{Aesthetic \\Score}$\uparrow$ & \makecell{Imaging \\Score}$\uparrow$ & CLIP-cap$\uparrow$ & Style-ref$\uparrow$ & \makecell{Aesthetic \\Score}$\uparrow$ & \makecell{Imaging\\ Score}$\uparrow$ & CLIP-cap$\uparrow$ & DINO-ref$\uparrow$ \\
        \midrule 
        ACE         & 5.352  & 54.953  & 0.265   & 0.329 & 5.312 & 58.960 & 0.116 & \textbf{0.802} & 5.228 & 55.748 & 0.249       & \textbf{0.878}     \\
        OmniGen     & \textbf{5.790}  & \textbf{72.667}  & \textbf{0.270}   & 0.573 & \textbf{5.785} & \textbf{70.827} & \textbf{0.215} & 0.432 & \textbf{5.821} & 71.355 & \textbf{0.266}       & 0.753     \\
        IP-Adapter  & 5.055  & 64.239  & 0.254   & \textbf{0.633} & 5.773 & 69.629 & 0.144 & 0.749 & 5.726 & 70.329 & 0.242       & 0.841     \\
        ACE++       & 5.508  & 67.900  & 0.261   & 0.506 & -     & -     & -     & -     & 5.198 & 62.751 & 0.238       & 0.852     \\
        OminiControl& -      & -      & -       & -     & -     & -     & -     & -     & 5.651 & \textbf{72.273} & 0.264       & 0.783     \\
        \bottomrule
      \end{tabular}
  }
  \label{tab:supp_eval_ref_cre_metric}
\end{table*}
\begin{table*}[ht]
  \centering
  \caption{Metrics on Global Editing Tasks (Tasks 5-16).}
  \vspace{-8pt}
  \footnotesize
  \setlength{\tabcolsep}{1.3pt}{

     %subtable1
    \resizebox{\textwidth}{!}{
      \begin{tabular}{c|cccccc|cccccc|cccccc}
        \toprule
        \multirow{2}{*}{Models} & \multicolumn{6}{c}{\textbf{Task 5: Color Editing}} & \multicolumn{6}{c}{\textbf{Task 6: Motion Editing}} & \multicolumn{6}{c}{\textbf{Task 7: Face Editing}} \\
        & \makecell{Aesthetic\\Score}$\uparrow$ & \makecell{Imaging\\Score}$\uparrow$ & CLIP-cap$\uparrow$ & VLLM-QA$\uparrow$ & CLIP-src$\uparrow$ & L1-src$\downarrow$ & \makecell{Aesthetic\\Score}$\uparrow$ & \makecell{Imaging\\Score}$\uparrow$ & CLIP-cap$\uparrow$ & VLLM-QA$\uparrow$ & CLIP-src$\uparrow$ & L1-src$\downarrow$ & \makecell{Aesthetic\\Score}$\uparrow$ & \makecell{Imaging\\Score}$\uparrow$ & CLIP-cap$\uparrow$ & VLLM-QA$\uparrow$ & CLIP-src$\uparrow$ & L1-src$\downarrow$ \\
        \midrule 
        ACE & \textbf{5.244} & 55.219 & \textbf{0.285} & \textbf{0.896} & \textbf{0.919} & \textbf{0.080} & \textbf{5.146} & 57.679 & \textbf{0.278} & \textbf{0.354} & \textbf{0.946} & \textbf{0.033} & 4.798 & 56.851 & \textbf{0.268} & \textbf{0.796} & \textbf{0.899} & \textbf{0.046} \\
        OmniGen & 4.918 & \textbf{63.562} & 0.277 & 0.789 & 0.880 & 0.119 & 4.927 & \textbf{61.038} & 0.262 & 0.329 & 0.870 & 0.106 & 4.735 & \textbf{63.584} & 0.247 & 0.636 & 0.818 & 0.095\\
        InstructPix2Pix & 4.990 & 53.124 & 0.267 & 0.452 & 0.828 & 0.217 & 4.796 & 57.453 & 0.211 & 0.081 & 0.719 & 0.134 & \textbf{4.920} & 57.941 & 0.192 & 0.364 & 0.669 & 0.151\\
        MagicBrush & 4.826 & 51.677 & 0.267 & 0.604 & 0.854 & 0.094 & 4.620 & 53.121 & 0.254 & 0.267 & 0.826 & 0.081 & 4.636 & 55.833 & 0.258 & 0.660 & 0.836 & 0.054\\
        UltraEdit & 5.136 & 52.398 & 0.274 & 0.485 & 0.864 & 0.098 & 4.970 & 55.514 & 0.266 & 0.199 & 0.871 & 0.059 & 4.774 & 57.159 & 0.247 & 0.655 & 0.786 & 0.057\\
        \bottomrule
      \end{tabular}
      }

       %subtable2
    \resizebox{\textwidth}{!}{
      \begin{tabular}{c|cccccc|cccccc|cccccc}
        \toprule
        \multirow{2}{*}{Models} & \multicolumn{6}{c}{\textbf{Task 8: Texture Editing}} & \multicolumn{6}{c}{\textbf{Task 9: Style Editing}} & \multicolumn{6}{c}{\textbf{Task 10: Scene Editing}} \\
        & \makecell{Aesthetic\\Score}$\uparrow$ & \makecell{Imaging\\Score}$\uparrow$ & CLIP-cap$\uparrow$ & VLLM-QA$\uparrow$ & CLIP-src$\uparrow$ & L1-src$\downarrow$ & \makecell{Aesthetic\\Score}$\uparrow$ & \makecell{Imaging\\Score}$\uparrow$ & CLIP-cap$\uparrow$ & VLLM-QA$\uparrow$ & CLIP-src$\uparrow$ & L1-src$\downarrow$ & \makecell{Aesthetic\\Score}$\uparrow$ & \makecell{Imaging\\Score}$\uparrow$ & CLIP-cap$\uparrow$ & VLLM-QA$\uparrow$ & CLIP-src$\uparrow$ & L1-src$\downarrow$ \\
        \midrule 
        ACE & \textbf{5.408} & 57.106 & \textbf{0.276} & 0.605 & \textbf{0.918} & \textbf{0.060} & \textbf{4.967} & 51.081 & \textbf{0.258} & 0.470 & \textbf{0.781} & 0.158 & 5.076 & 47.345 & \textbf{0.253} & 0.392 & \textbf{0.902} & \textbf{0.075} \\
        OmniGen & 5.151 & \textbf{64.069} & 0.257 & 0.558 & 0.819 & 0.156 & 4.935 & \textbf{60.567} & 0.250 & \textbf{0.478} & 0.763 & 0.183 & \textbf{5.109} & \textbf{55.674} & 0.246 & 0.414 & 0.806 & 0.169\\
        InstructPix2Pix & 4.847 & 59.220 & 0.240 & 0.422 & 0.703 & 0.193 & 4.630 & 48.674 & 0.228 & 0.416 & 0.627 & 0.218 & 5.048 & 45.324 & 0.224 & 0.381 & 0.657 & 0.219\\
        MagicBrush & 4.720 & 52.909 & 0.245 & 0.463 & 0.796 & 0.122 & 4.227 & 46.647 & 0.184 & 0.140 & 0.600 & 0.249 & 4.592 & 44.262 & 0.239 & \textbf{0.464} & 0.725 & 0.189\\
        UltraEdit & 5.148 & 54.875 & 0.270 & \textbf{0.714} & 0.821 & 0.093 & 4.697 & 49.067 & 0.246 & 0.414 & 0.726 & \textbf{0.093} & 5.023 & 44.961 & 0.255 & 0.453 & 0.764 & 0.098\\
        \bottomrule
      \end{tabular}
      }

      %subtable3
      \resizebox{\textwidth}{!}{
      \begin{tabular}{c|cccccc|cccccc|cccccc}
        \toprule
        \multirow{2}{*}{Models} & \multicolumn{6}{c}{\textbf{Task 11: Subject Addition}} & \multicolumn{6}{c}{\textbf{Task 12: Subject Removal}} & \multicolumn{6}{c}{\textbf{Task 13: Subject Change}} \\
        & \makecell{Aesthetic\\Score}$\uparrow$ & \makecell{Imaging\\Score}$\uparrow$ & CLIP-cap$\uparrow$ & VLLM-QA$\uparrow$ & CLIP-src$\uparrow$ & L1-src$\downarrow$ & \makecell{Aesthetic\\Score}$\uparrow$ & \makecell{Imaging\\Score}$\uparrow$ & CLIP-cap$\uparrow$ & VLLM-QA$\uparrow$ & CLIP-src$\uparrow$ & L1-src$\downarrow$ & \makecell{Aesthetic\\Score}$\uparrow$ & \makecell{Imaging\\Score}$\uparrow$ & CLIP-cap$\uparrow$ & VLLM-QA$\uparrow$ & CLIP-src$\uparrow$ & L1-src$\downarrow$ \\
        \midrule 
        ACE & 4.920 & 50.514 & \textbf{0.274} & \textbf{0.619} & \textbf{0.888} & \textbf{0.045} & 4.877 & 45.559 & \textbf{0.253} & \textbf{0.834} & 0.855 & \textbf{0.053} & \textbf{5.018} & 52.386 & \textbf{0.274} & 0.500 & \textbf{0.881} & \textbf{0.070}\\
        OmniGen & \textbf{4.987} & \textbf{58.151} & 0.266 & 0.611 & 0.877 & 0.077 & 4.884 & \textbf{54.001} & 0.231 & 0.611 & 0.830 & 0.107 & 4.997 & \textbf{59.282} & 0.262 & 0.460 & 0.812 & 0.115\\
        InstructPix2Pix & 4.884 & 52.320 & 0.205 & 0.234 & 0.703 & 0.144 & 4.827 & 48.625 & 0.170 & 0.119 & 0.711 & 0.141 & 4.746 & 53.884 & 0.229 & 0.360 & 0.691 & 0.179\\
        MagicBrush & 4.656 & 46.127 & 0.272 & 0.594 & 0.866 & 0.061 & 4.672 & 45.197 & 0.231 & 0.322 & 0.864 & 0.069 & 4.291 & 48.950 & 0.257 & 0.500 & 0.756 & 0.123\\
        UltraEdit & 4.932 & 47.651 & 0.259 & 0.537 & 0.830 & 0.064 & \textbf{4.974} & 47.308 & 0.223 & 0.256 & \textbf{0.873} & 0.056 & 4.868 & 51.984 & 0.269 & \textbf{0.540} & 0.788 & 0.082\\
        \bottomrule
      \end{tabular}
      }

      %subtable4
      \resizebox{\textwidth}{!}{
      \begin{tabular}{c|cccccc|cccccc|cccccc}
        \toprule
        \multirow{2}{*}{Models} & \multicolumn{6}{c}{\textbf{Task 14: Text Render}} & \multicolumn{6}{c}{\textbf{Task 15: Text Removal}} & \multicolumn{6}{c}{\textbf{Task 16: Composite Editing}} \\
        & \makecell{Aesthetic\\Score}$\uparrow$ & \makecell{Imaging\\Score}$\uparrow$ & CLIP-cap$\uparrow$ & VLLM-QA$\uparrow$ & CLIP-src$\uparrow$ & L1-src$\downarrow$ & \makecell{Aesthetic\\Score}$\uparrow$ & \makecell{Imaging\\Score}$\uparrow$ & CLIP-cap$\uparrow$ & VLLM-QA$\uparrow$ & CLIP-src$\uparrow$ & L1-src$\downarrow$ & \makecell{Aesthetic\\Score}$\uparrow$ & \makecell{Imaging\\Score}$\uparrow$ & CLIP-cap$\uparrow$ & VLLM-QA$\uparrow$ & CLIP-src$\uparrow$ & L1-src$\downarrow$ \\
        \midrule 
        ACE & 3.981 & 51.104 & \textbf{0.263} & 0.517 & 0.800 & \textbf{0.052} & \textbf{4.842} & 49.714 & \textbf{0.270} & \textbf{0.754} & \textbf{0.883} & \textbf{0.037} & \textbf{5.475} & 49.984& 0.270 & 0.420 & \textbf{0.797} & 0.194 \\
        OmniGen & 4.351 & \textbf{57.420} & \textbf{0.263} & \textbf{0.596} & 0.815 & 0.075 & 4.500 & \textbf{57.211} & 0.223 & 0.330 & 0.767 & 0.125 & 5.259 & \textbf{62.885} & 0.272 & \textbf{0.567} & 0.753 & 0.229 \\
        InstructPix2Pix & \textbf{4.712} & 51.201 & 0.213 & 0.010 & 0.718 & 0.187 & 4.400 & 44.069 & 0.194 & 0.147 & 0.655 & 0.163 & 4.827 & 50.006 & 0.258 & 0.280 & 0.698 & \textbf{0.237}  \\
        MagicBrush & 4.458 & 45.903 & 0.261 & 0.099 & \textbf{0.845} & 0.088 & 4.359 & 44.484 & 0.260 & 0.529 & 0.838 & 0.063 & 4.665 & 47.646 & 0.245 & 0.070 & 0.732 & 0.185 \\
        UltraEdit & 4.465 &	46.965 & 0.262 & 0.187 & 0.813 & 0.059 & 4.640 & 47.908 & 0.255 & 0.246 & 0.861 & 0.044 & 5.180 & 48.372 & \textbf{0.274} & 0.395 & 0.731 & 0.147 \\
        \bottomrule
      \end{tabular}
      }
  }
  \label{tab:supp_eval_noref_edit_metric_global}
\end{table*}
\begin{table*}[ht]
  \centering
  \caption{Metrics on Local Editing Tasks (Tasks 17-22).}
  \vspace{-8pt}
  \footnotesize
  \setlength{\tabcolsep}{1.3pt}{
     %subtable1
    \resizebox{\textwidth}{!}{
      \begin{tabular}{c|cccccc|cccccc|cccccc}
        \toprule
        \multirow{2}{*}{Models} & \multicolumn{6}{c}{\textbf{Task 17: Inpainting}} & \multicolumn{6}{c}{\textbf{Task 18: Outpainting}} & \multicolumn{6}{c}{\textbf{Task 19: Local Subject Addition}} \\
        & \makecell{Aesthetic\\Score}$\uparrow$ & \makecell{Imaging\\Score}$\uparrow$ & CLIP-cap$\uparrow$ & VLLM-QA$\uparrow$ & CLIP-src$\uparrow$ & L1-src$\downarrow$ & \makecell{Aesthetic\\Score}$\uparrow$ & \makecell{Imaging\\Score}$\uparrow$ & CLIP-cap$\uparrow$ & VLLM-QA$\uparrow$ & CLIP-src$\uparrow$ & L1-src$\downarrow$ & \makecell{Aesthetic\\Score}$\uparrow$ & \makecell{Imaging\\Score}$\uparrow$ & CLIP-cap$\uparrow$ & VLLM-QA$\uparrow$ & CLIP-src$\uparrow$ & L1-src$\downarrow$ \\
        \midrule 
        ACE & 4.878 & 51.793 & 0.269 & 0.833 & 0.785 & 0.024 & 5.514 & 50.403 & 0.287 & 0.376 & 0.891 & 0.017 & 4.965 & 51.704 & 0.272 & 0.555 & 0.897 & 0.029  \\
        OmniGen & 4.545 & 59.264 & 0.238 & 0.524 & 0.734 & 0.108 & 5.442 &	\textbf{65.758} & 0.265 & 0.326 & 0.802 & 0.114 & 4.584 & 58.911 & 0.249 & 0.479 & 0.814 & 0.066  \\
        ACE++ & \textbf{5.064} & \textbf{61.661} & \textbf{0.272} & \textbf{0.910} & 0.776 & \textbf{0.016} & \textbf{5.644} & 64.156 & \textbf{0.289} & \textbf{0.531} & 0.908 & \textbf{0.010} & \textbf{5.014} & \textbf{62.083} & 0.268 & \textbf{0.785} & 0.894 & \textbf{0.018} \\
        UltraEdit & 3.817 &	46.284 & 0.250 & 0.180 & \textbf{0.952} & 0.019 & 4.498 & 43.968 & 0.274 & 0.220 & \textbf{0.945} & 0.018 & 4.881 & 47.855 & \textbf{0.275} & 0.555 & \textbf{0.909} & 0.021 \\
        \bottomrule
      \end{tabular}
      }

       %subtable2
    \resizebox{\textwidth}{!}{
      \begin{tabular}{c|cccccc|cccccc|cccccc}
        \toprule
        \multirow{2}{*}{Models} & \multicolumn{6}{c}{\textbf{Task 20: Local Subject Removal}} & \multicolumn{6}{c}{\textbf{Task 21: Local Text Render}} & \multicolumn{6}{c}{\textbf{Task 22: Local Text Removal}} \\
        & \makecell{Aesthetic\\Score}$\uparrow$ & \makecell{Imaging\\Score}$\uparrow$ & CLIP-cap$\uparrow$ & VLLM-QA$\uparrow$ & CLIP-src$\uparrow$ & L1-src$\downarrow$ & \makecell{Aesthetic\\Score}$\uparrow$ & \makecell{Imaging\\Score}$\uparrow$ & CLIP-cap$\uparrow$ & VLLM-QA$\uparrow$ & CLIP-src$\uparrow$ & L1-src$\downarrow$ & \makecell{Aesthetic\\Score}$\uparrow$ & \makecell{Imaging\\Score}$\uparrow$ & CLIP-cap$\uparrow$ & VLLM-QA$\uparrow$ & CLIP-src$\uparrow$ & L1-src$\downarrow$ \\
        \midrule 
        ACE & 4.996 & 47.011 & \textbf{0.258} & \textbf{0.757} & 0.852 & 0.024 & 4.275 &	43.159 & 0.276 & \textbf{0.791} &	0.860 &	0.016 & \textbf{4.896} &	49.766 & \textbf{0.273} &	\textbf{0.801} &	0.888 &	0.033 \\
        OmniGen & 4.792 & 54.320 & 0.238 & 0.658 & 0.787 & 0.061 & 4.015 & 42.527 & 0.261 & 0.380 & 0.815 & 0.066 & 4.487 & 56.398 & 0.246 & 0.674 & 0.793 & 0.097  \\
        ACE++ & \textbf{5.061} & \textbf{61.614} & 0.229 & 0.312 & \textbf{0.901} &	\textbf{0.017} & 4.231 &	\textbf{43.276} & \textbf{0.277} & 0.834 &	0.899 &	\textbf{0.012} & 4.694 &	\textbf{59.636} & 0.260 &	0.704 &	0.905 &	\textbf{0.017}  \\
        UltraEdit & 4.858 &	48.748 & 0.226 & 0.287 &	0.888 &	0.018 & \textbf{4.506} &	38.887 & \textbf{0.277} &	0.098 &	\textbf{0.946} &	0.014 & 4.665 & 47.294 & 0.264 & 0.714 & \textbf{0.910} & 0.023  \\
        \bottomrule
      \end{tabular}
      }
      }
  \label{tab:supp_eval_noref_edit_metric_local}
\end{table*}

\begin{figure*}[ht]
  \centering
  \includegraphics[width=\linewidth]{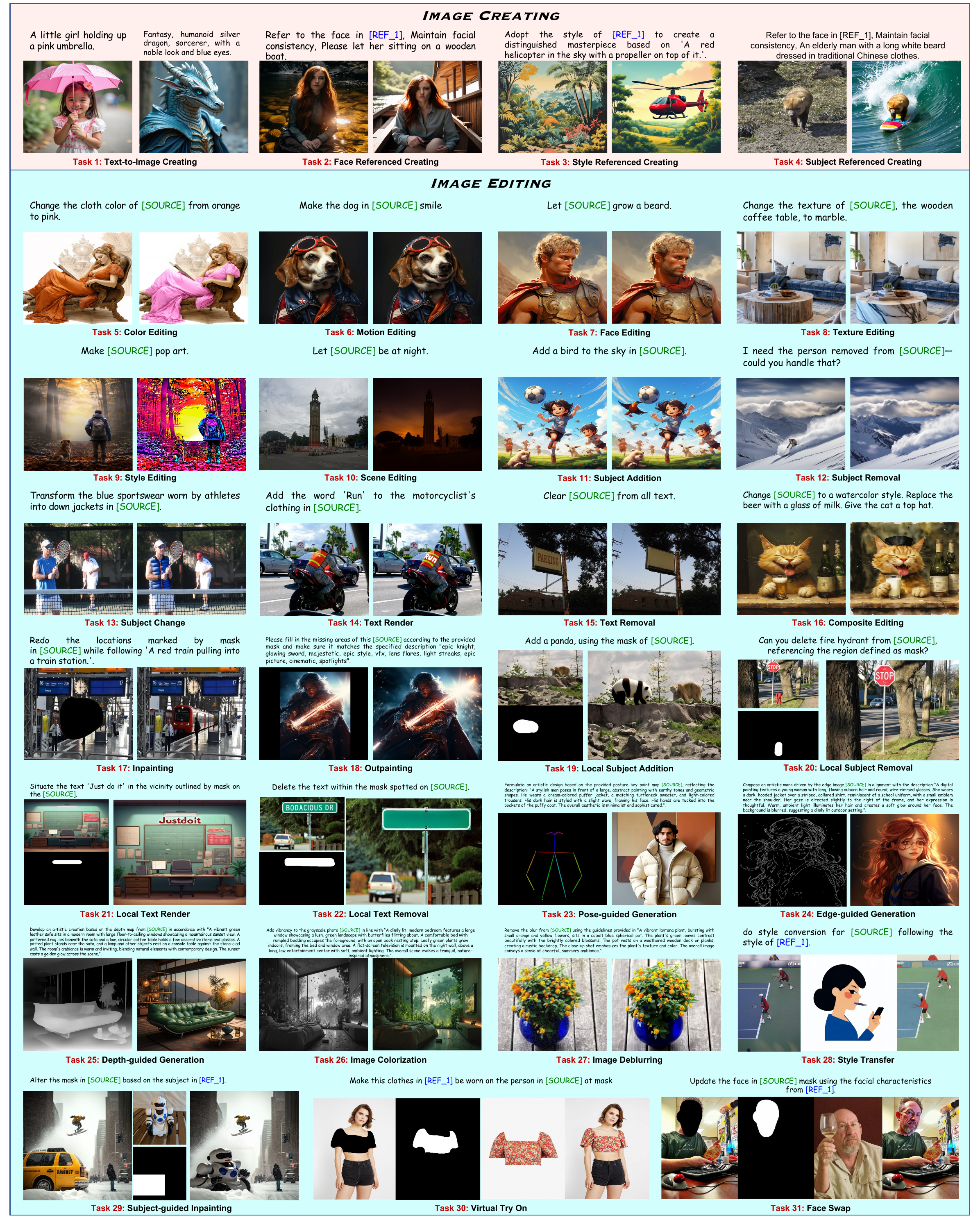}
   \caption{Examples of 31 fine-grained evaluation tasks in our ICE-Bench.}
   \label{fig:evatask}
\end{figure*}

\end{document}